\documentclass[10pt,twocolumn,letterpaper]{article}

\usepackage{iccv}
\usepackage{times}
\usepackage{epsfig}
\usepackage{graphicx}
\usepackage{amsmath}
\usepackage{amssymb}
\usepackage{subfig}
\usepackage[dvipsnames]{xcolor}
\usepackage{floatrow}
\usepackage{adjustbox}
\usepackage{caption}
\newif\ifsqueeze
\squeezetrue  % set \squeezetrue or \squeezefalse to enable or disable squeezing
\ifsqueeze
  
  \newcommand{\Section}[1]{\vspace{-1mm} \section{#1} \vspace{-2mm}}
  
  \newcommand{\Subsection}[1]{\vspace{-1mm} \subsection{#1} \vspace{-1mm} }
  \newcommand{\Subsubsection}[1]{\vspace{0mm} \subsubsection{#1} \vspace{-2mm} }
  
\else
  
  \newcommand{\Section}[1]{\section{#1}}
  
  \newcommand{\Subsection}[1]{\subsection{#1}}
  \newcommand{\Subsubsection}[1]{\subsubsection{#1}}

\fi

\usepackage[pagebackref=true,breaklinks=true,letterpaper=true,colorlinks,bookmarks=false]{hyperref}

\iccvfinalcopy % *** Uncomment this line for the final submission

\def\iccvPaperID{3094} % *** Enter the ICCV Paper ID here

\ificcvfinal\fi
\begin{document}

%%%%%%%%% TITLE
%\title{Learning the long tail from synthetic and real data simultaneously}
\title{Synthetic Examples Improve Generalization for Rare Classes}
\maketitle
\thispagestyle{empty}
%%%%%%%%% ABSTRACT
\begin{abstract}
   The ability to detect and classify rare occurrences in images has important applications -- for example, counting rare and endangered species when studying biodiversity, or detecting infrequent traffic scenarios that pose a danger to self-driving cars. Few-shot learning is an open problem: current computer vision systems struggle to categorize objects they have seen only rarely during training, and collecting a sufficient number of training examples of rare events is often challenging and expensive, and sometimes outright impossible. We explore in depth an approach to this problem: complementing the few available training images with ad-hoc simulated data. 
   
   Our testbed is animal species classification, which has a real-world long-tailed distribution.  We analyze the effect of different axes of variation in simulation, such as pose, lighting, model, and simulation method, and we prescribe best practices for efficiently incorporating simulated data for real-world performance gain. Our experiments reveal that synthetic data can considerably reduce error rates for classes that are rare, that as the amount of simulated data is increased, accuracy on the target class improves, and that high variation of simulated data provides maximum performance gain. 
\end{abstract}

%%%%%%%%% BODY TEXT
\Section{Introduction}
\begin{figure*}[!ht]
  \centering
  \subfloat{\includegraphics[width=3cm]{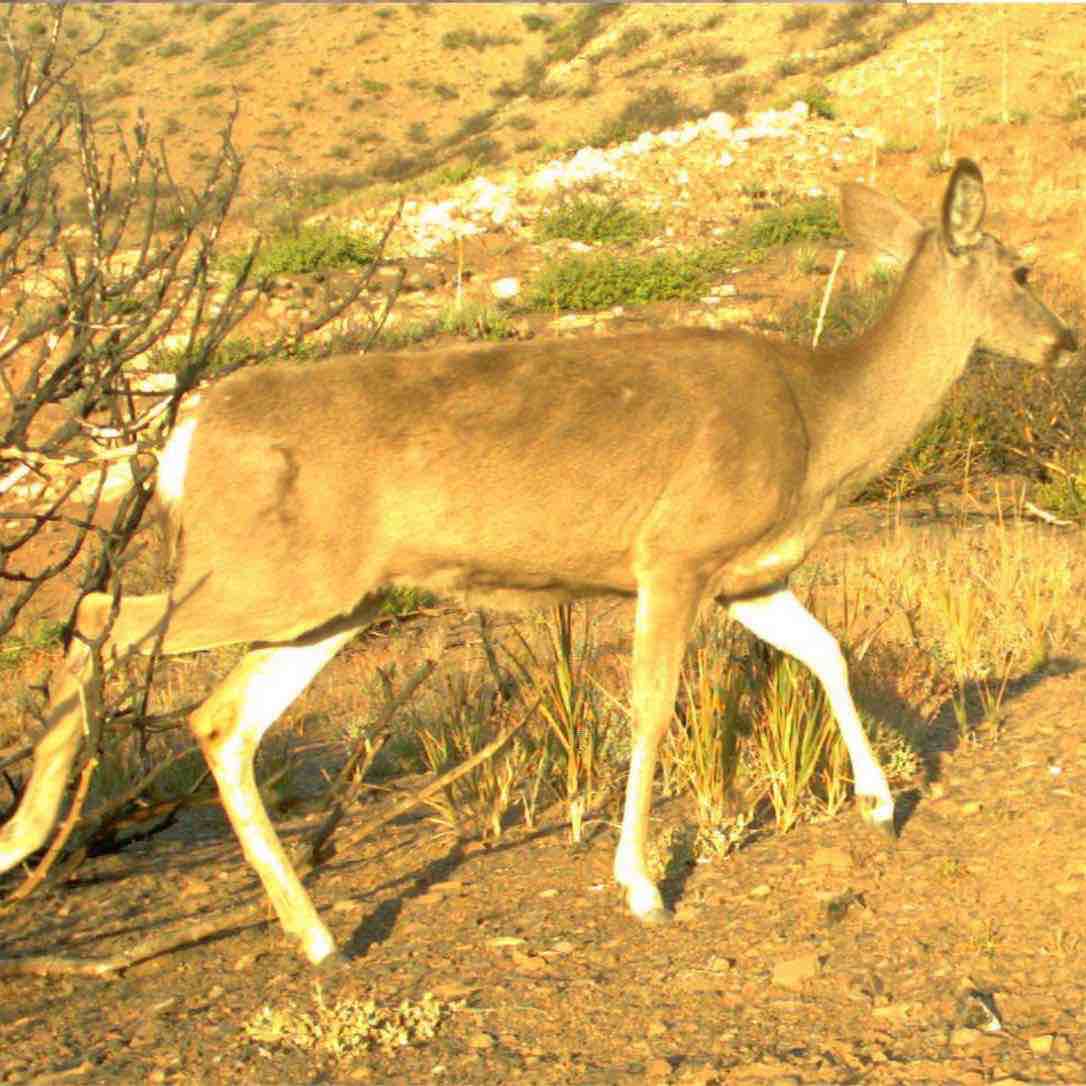}}\quad
    \subfloat{\includegraphics[width=3cm]{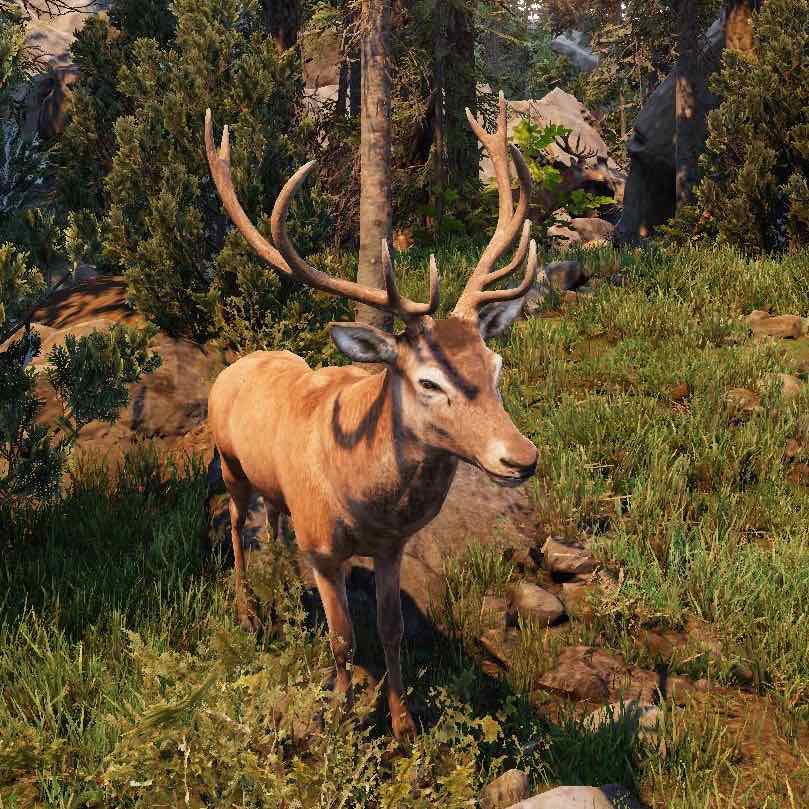}}\quad
        \subfloat{\includegraphics[width=3cm]{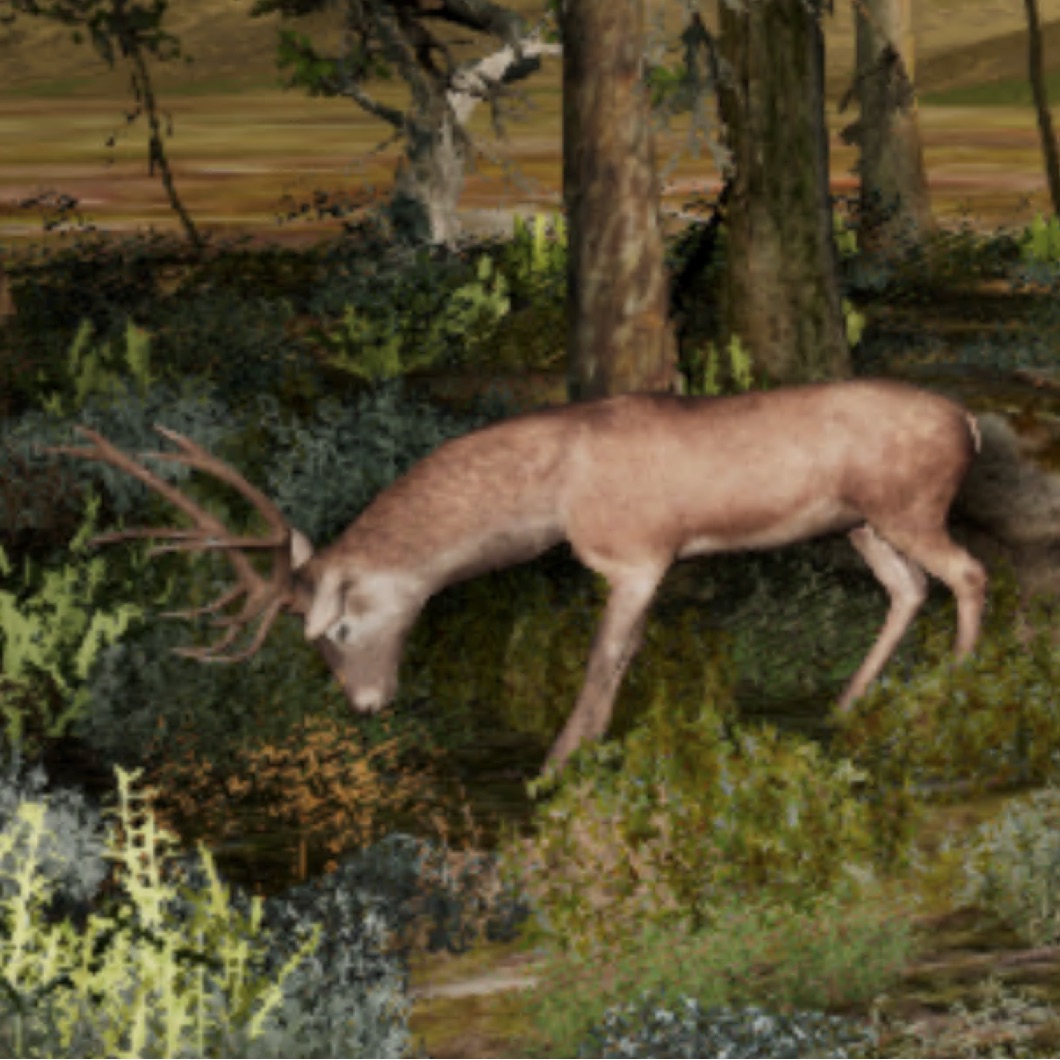}}\quad
 \subfloat{\includegraphics[width=3cm]{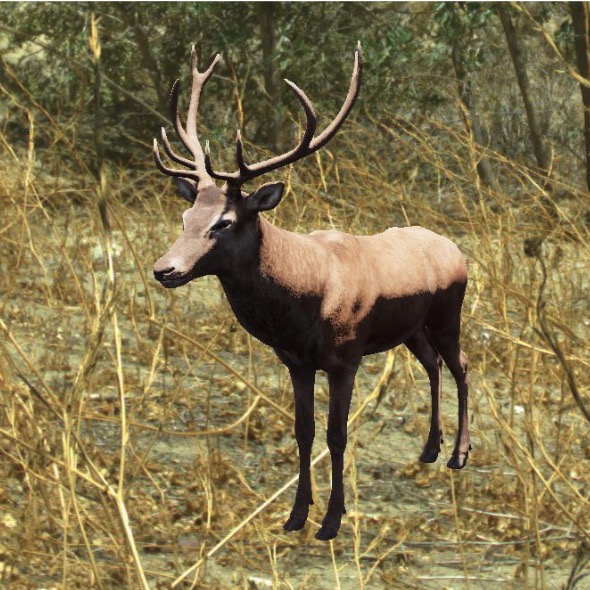}}\quad
 \subfloat{\includegraphics[width=3cm]{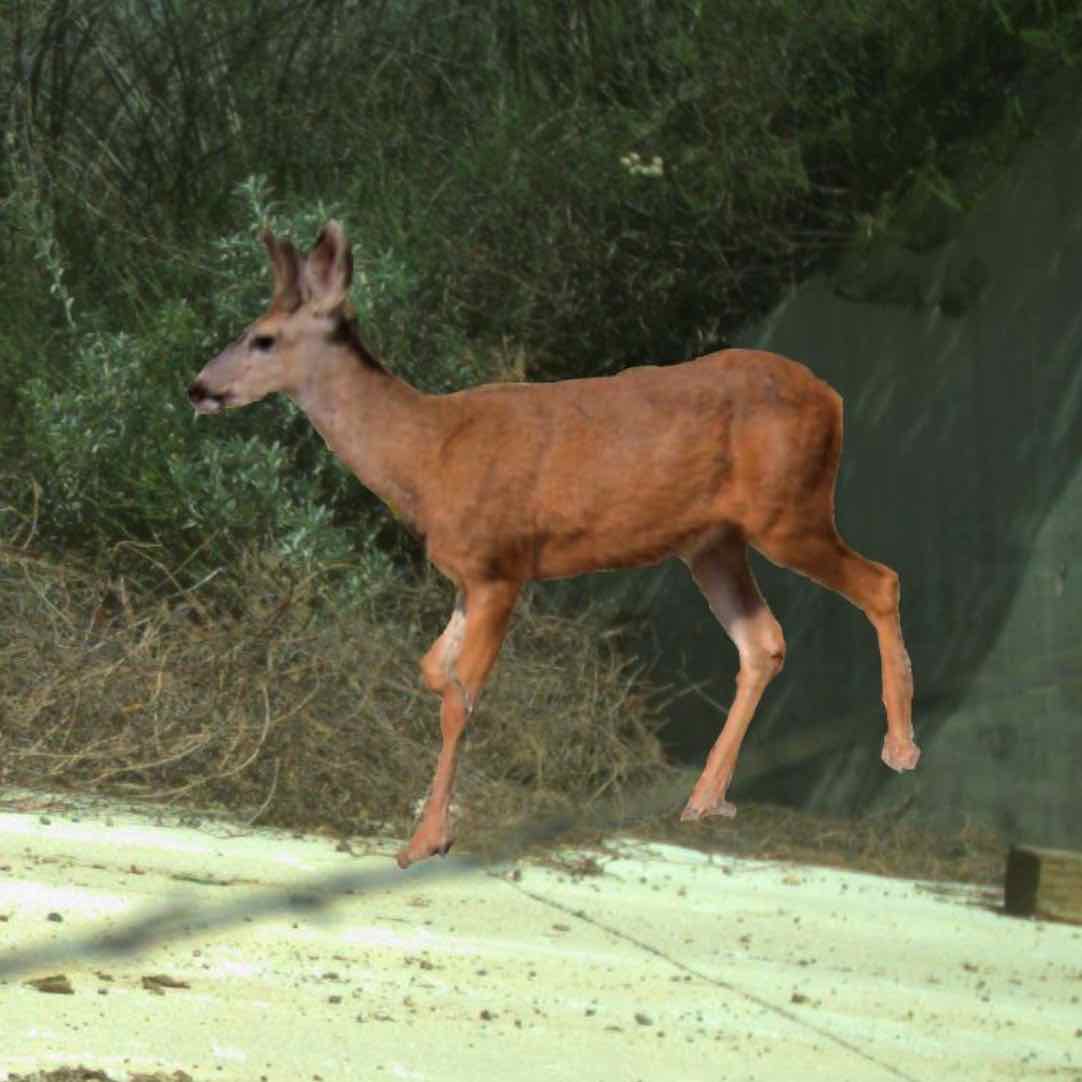}}\\
    
    \subfloat[Real Camera Traps]{\includegraphics[width=3cm]{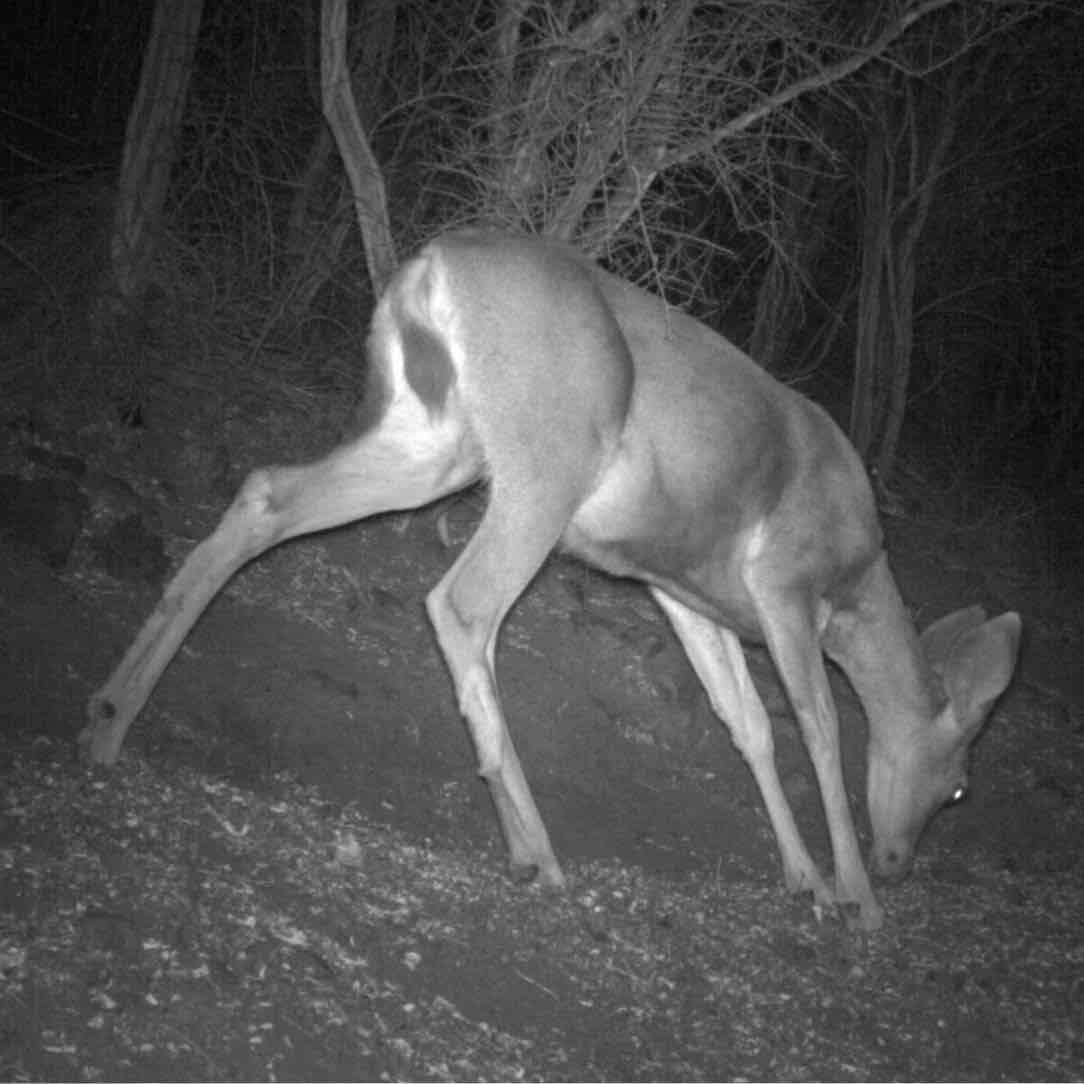}}\quad
  \subfloat[TrapCam-Unity]{\includegraphics[width=3cm]{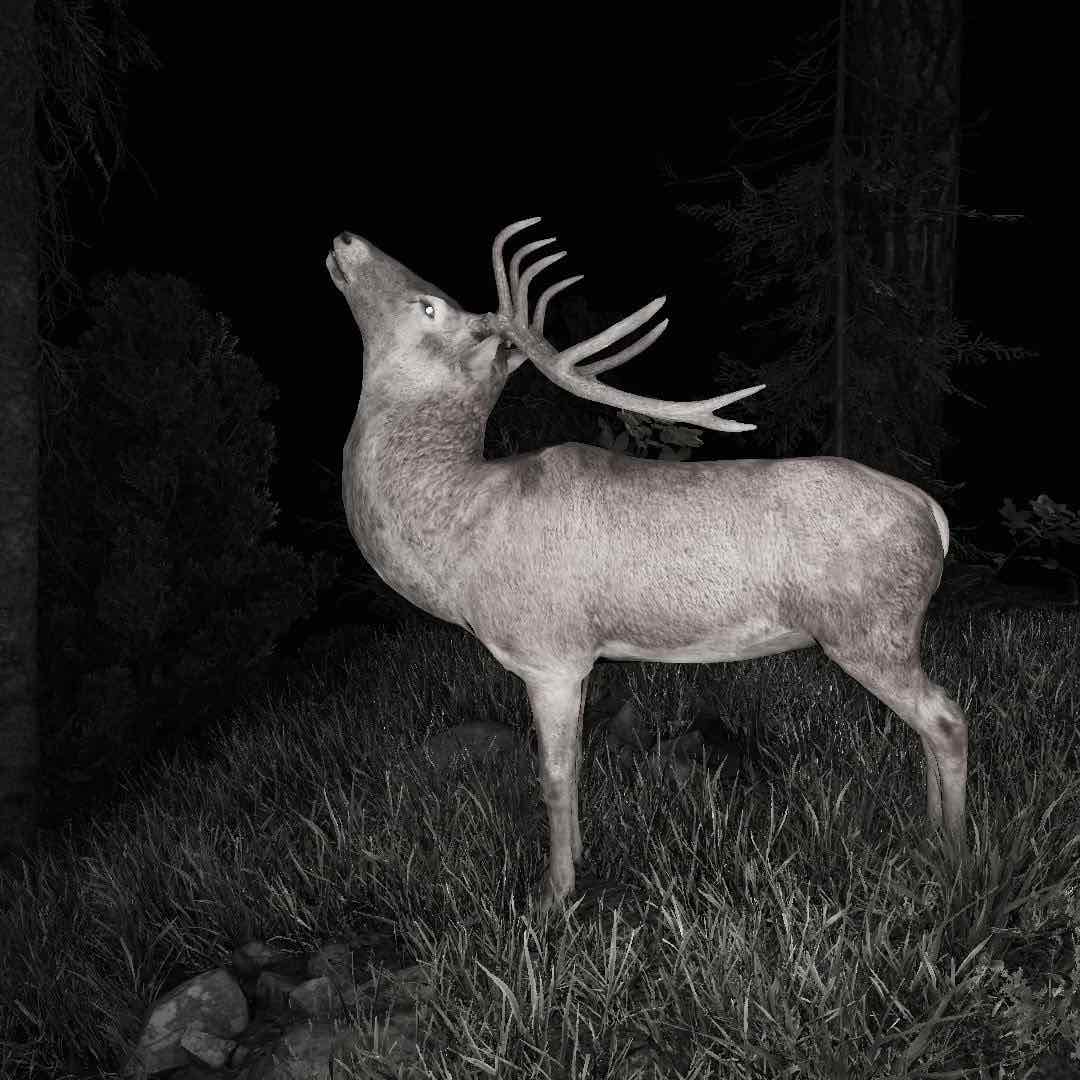}}\quad
        \subfloat[TrapCam-AirSim]{\includegraphics[width=3cm]{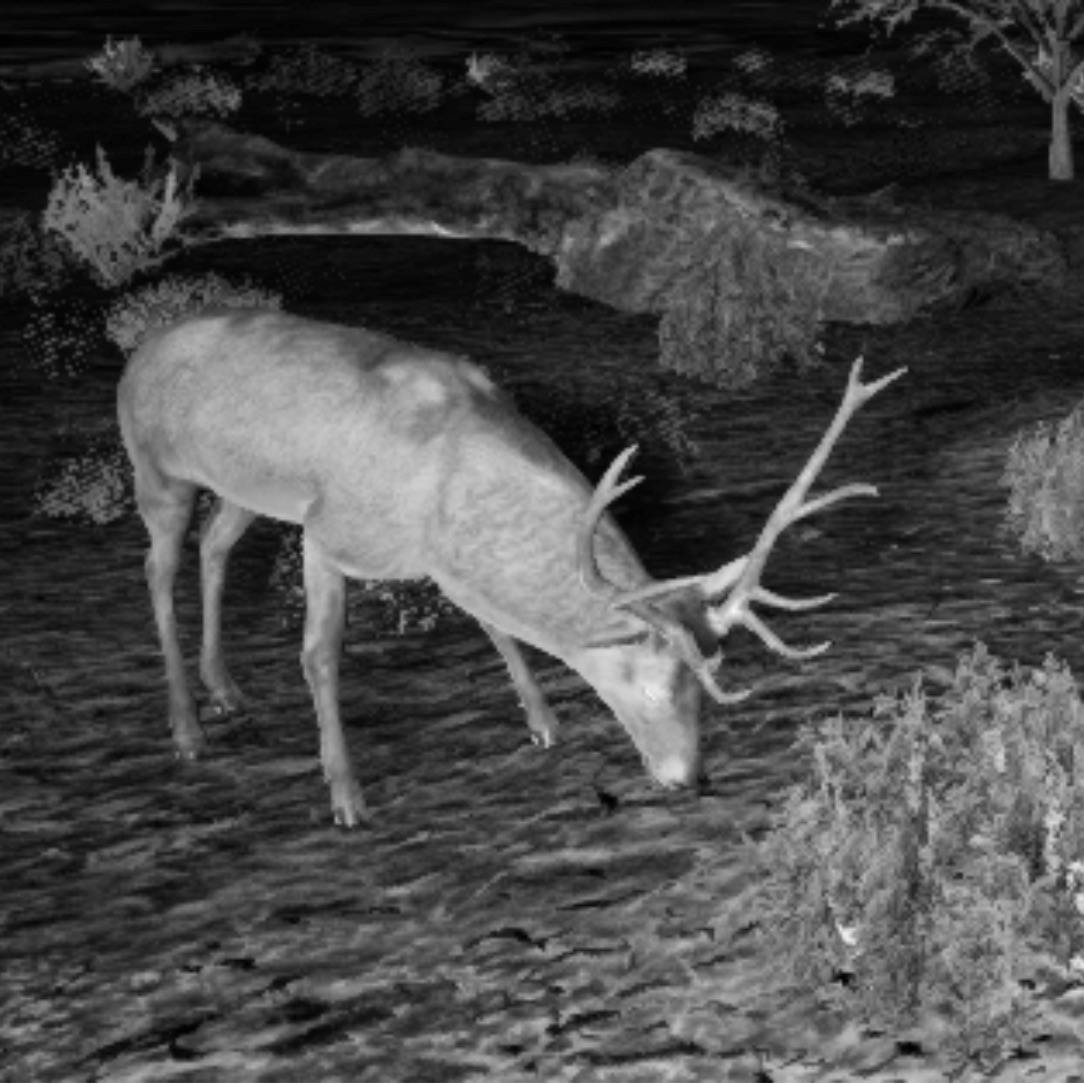}}\quad
          \subfloat[Sim on Empty]{\includegraphics[width=3cm]{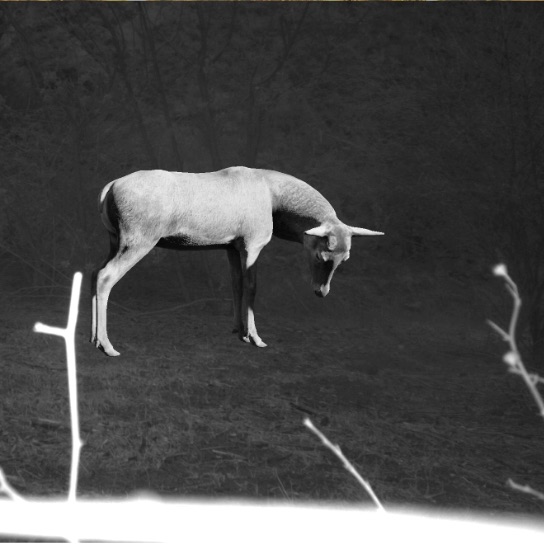}}\quad
  \subfloat[Real on Empty]{\includegraphics[width=3cm]{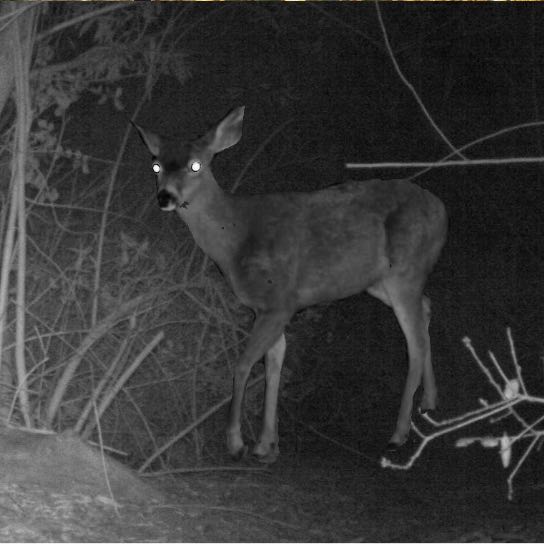}}
\vspace{-5pt}
\setlength{\belowcaptionskip}{-25pt}
  \caption{\textbf{Day and night examples for each simulation method.} We compare four different simulation methods and compare the effects of each on classification performance.}
  \label{fig:night}
  \label{fig:sim_methods}
\end{figure*}

In recent years computer vision researchers have made substantial progress towards automated visual recognition across a wide variety of visual domains~\cite{russakovsky2015imagenet,esteva2017dermatologist,poplin2018prediction,van2017inaturalist,norouzzadeh2017automatically,merlinBirdID,Beery_2018_ECCV}.  However, applications are hampered by the fact that in the real world the distribution of visual classes is long-tailed, and state-of-the-art recognition algorithms struggle to learn classes with limited data ~\cite{van2017devil}. In some cases (such as recognition of rare endangered species) classifying rare occurrences correctly  is crucial. Simulated data, which is plentiful, and comes with annotation ``for free", has been shown to be useful for various computer vision tasks \cite{Varol_2017_CVPR, pepik2015holding,hinterstoisser2019annotation,rajpura2017object, goodfellow2014generative,Shrivastava_2017_CVPR,Richter_2016,peng2018syn2real,hattori2015learning}. However, an exploration of this approach in a long-tailed setting is still missing (see Section \ref{datasimulation}).

As a testbed, we focus on the effect of simulated data augmentation on the real-world application of recognizing animal species in camera trap images. Camera traps are heat- or motion-activated cameras placed in the wild to monitor animal populations and behavior. The processing of camera trap images is currently limited by human review capacity; consequently, automated detection and classification of animals is a necessity for scalable biodiversity assessment.  A single sighting of a rare species is of immense importance. However, training data of rare species is, by definition, scarce. This makes this domain ideal for studying methods for training detection and classification algorithms with few training examples. We utilize a technique from \cite{Beery_2018_ECCV} which tests performance at camera locations both seen (cis) and unseen (trans) during training in order to explicitly study generalization (see Section \ref{data} for a more detailed explanation).

We investigate the use of simulated data as augmentation during training, and how to best combine real data for common classes with simulated data for rare classes to achieve optimal performance across the class set at test time. We consider four different data simulation methods (see Fig.\ref{fig:sim_methods}) and compare the effects of each on classification performance. Finally, we analyze the effect of both increasing the number of simulated images and controlling for axes of variation to provide best practices for leveraging simulated data for real-world performance gain on rare classes.

\Section{Related work}
\Subsection{Visual Categorization Datasets}
%Recently, methods for classification and detection have provided large performance improvements over traditional vision approaches~\cite{szegedy2016rethinking,huang2017speed,he2016deep}.
Large and well-annotated public datasets allow scientists to train, analyze, and compare the performance of different methods, and have provided large performance improvements over traditional vision approaches ~\cite{szegedy2016rethinking,huang2017speed,he2016deep}. The most popular datasets used for this purpose are ImageNet, COCO, PascalVOC, and OpenImages, all of which are human-curated from images scraped from the web~\cite{imagenet_cvpr09,lin2014microsoft,everingham2010pascal,openimages}. These datasets cover a wide set of classes across both the manufactured and natural world, and are usually designed to provide ``enough" data per class to avoid the low-data regime. More recently researchers have proposed datasets that focus specifically on the natural world, which has a long-tailed distribution \cite{van2017inaturalist,Beery_2018_ECCV,leafsnap_eccv2012}. The iNaturalist 2018 dataset \cite{van2017inaturalist} encourages a focus on the long tail by including classes with little training data. Caltech Camera Traps \cite{Beery_2018_ECCV} additionally introduced the challenge of learning in few locations, against constant backgrounds, and generalizing to new locations.

\Subsection{Handling Imbalanced Datasets}
Imbalanced datasets often lead to bias in algorithm performance toward well-represented classes \cite{buda2018systematic}. 
Algorithmic solutions often use a non-uniform cost per misclassification (called cost-sensitive learning~\cite{elkan2001foundations,he2008learning,he2008adasyn}) which encourage models to `focus' on training examples from rare classes. The simplest version of this uses weighted loss, where each incorrect example incurs loss inversely proportional to the number of representatives of that class~\cite{he2008learning}. For example, focal loss was recently proposed as a method for dealing with the large class imbalance innate to detection, where the majority of examples come from the background class~\cite{lin2018focal}.

Data solutions employ data augmentation, either by 1) over-sampling the minority classes, 2) under-sampling the majority classes, or 3) generating new examples for the minority classes. When using mini-batch gradient descent, oversampling the minority classes is similar to weighting these classes higher than the majority classes, as in cost-sensitive learning. Under-sampling the majority classes is non-ideal, as this may lose information about common classes. Our paper falls into the third category: augmenting the training data for rare classes. Data augmentation via pre-processing, using affine and photometric transformations, is a well-established tool for improving generalization \cite{krizhevsky2012imagenet, howard2013some}. Data generation and simulation have begun to be explored as data augmentation methods, see Section \ref{datasimulation}.

Algorithmic and data solutions for imbalanced data are complementary, algorithmic advances can be used in conjunction with augmented training data.

\Subsection{Low-shot Learning}
Low-shot learning attempts to learn categories from few examples \cite{li2006one}. Wang and Herbert \cite{wang2016learning} learn to classify with small amounts of training data by regressing from small-dataset classifiers to large-dataset classifiers. Hariharan and Girshick \cite{hariharan2017low} look specifically at ImageNet, using classes that are unbalanced, some with large amounts of training data, and some with little training data. Their proposed solution was beneficial within low-capacity models, but matched the performance of high-capacity models off-the-shelf. Metric learning learns a representation space where distance corresponds to similarity, and uses this as a basis for low-shot solutions \cite{cui2016fine}. 
We consider the low-shot regime with regard to \textit{real} data for our rare target class, but investigate the use of added synthetic data based on a human-generated articulated model of the unseen class during training instead of additional class-specific attribute labels at training and test time. This takes us outside of the traditional low-shot framework into the realm of domain transfer from simulated to real data. 

\Subsection{Data Augmentation via Style Transfer, Generation, and Simulation} \label{datasimulation}
Image generation via generative adversarial networks (GANs) and recurrent neural networks (RNNs), as well as style transfer and image-to-image translation have all been considered as sources for data augmentation \cite{bousmalis2017unsupervised,gregor2015draw,im2016generating, radford2015unsupervised, tran2017bayesian,luan2017deep, zhu2017unpaired}.
These techniques are valuable, but require large amounts of data to generate realistic images making them un-ideal solutions for low-data regimes. Though conditional generation allows for class-specific output, the results can be difficult to interpret or control.

Graphics engines such an Unreal \cite{UNREAL,qiu2017unrealcv} and Unity \cite{Unity} leverage the expertise of human artists and complex physics models to generate photorealistic simulated images, which can be used for data augmentation. Because ground truth is known at generation, simulated data has proved particularly useful for tasks requiring detailed and expensive annotation, such as keypoints, semantic segmentations, or depth information \cite{Varol_2017_CVPR, pepik2015holding,hinterstoisser2019annotation,rajpura2017object, goodfellow2014generative,Shrivastava_2017_CVPR,Richter_2016,peng2018syn2real,hattori2015learning}. Varol \etal \cite{Varol_2017_CVPR} use synthetically-generated humans placed on top of real image backgrounds as pretraining for human pose estimation, and find they get best results when fine-tuning a synthetically-trained model on limited real data. \cite{Shrivastava_2017_CVPR} uses a combination of unlabeled real data and labeled simulated data of the same class to improve real-world performance on an eye-tracking task by using GANs \cite{goodfellow2014generative} to improve the ``realness" of their synthetic data. This method requires a large number of unlabled examples from the target class.
% \cite{hattori2015learning} use geometrically consistent computer generated data for a fixed location in order to learn to detect pedestrians at that location, with no real training data. 
\cite{pepik2015holding,hinterstoisser2019annotation,rajpura2017object} find that simulated data improves detection performance, and the degree of realism and variability of simulation affects the amount of improvement. They consider only small sets of non-deformable man-made objects. 
Richter \etal~\cite{Richter_2016} %generated images using the Grand Theft Auto V video game,and 
showed that a segmentation model for city scenes trained with a subset of their real dataset and a synthetic set outperforms a model trained with the full real dataset. \cite{peng2018syn2real} proposes a dataset and benchmark for evaluating models for unsupervised domain transfer from synthetic to real data with all-simulated training data, as opposed to simulated data only for classes with little representation. 
While this literature is encouraging, a number of questions are left unexplored. The first is a careful analysis of when simulated data is useful and, in particular, if it is useful in generalizing to new scenarios. Second, whether simulated data can be useful in highly complex and relatively unpredictable scenes such as natural scenes, as opposed to indoors and urban scenes. Third, whether it is just the synthetic objects or also the synthetic environments that contribute to learning.

\Subsection{Simulated Datasets}
Previous efforts on synthetic dataset generation focus on non-deformable man-made objects and indoor scenes \cite{song2016ssc, savva2017minos,wu2018building,hinterstoisser2019annotation,rajpura2017object,DBLP:journals/corr/abs-1712-05474}, human pose/actions \cite{Varol_2017_CVPR, de2017procedural}, or urban scenes \cite{ros2016synthia,gaidon2016virtual,Richter_2016,Dosovitskiy17}.

Bondi \etal~\cite{bondi2018airsim} previously released the AirSim-w data simulator within the domain of wildlife conservation, but it is focused on creating aerial infrared imagery. The resolution and quality of the assets is sufficient to replicate data from $100$ meters in the air, but is not realistic close-up. We contribute the first image data generators specifically for the natural world with the ability to recreate natural environments and generate near-photorealistic images of animals with real-world nuisance factors such as challenging pose, lighting, and occlusion within the scene. Our generators use high-quality 3D animated models to create realistic natural scenes at a depth of as little as one meter. 

\Section{Data and Simulation}
\Subsection{Real Data} \label{data}
Our real-world training and test data comes from the Caltech Camera Traps (CCT) dataset~\cite{Beery_2018_ECCV}. CCT contains $243,187$ images from $140$ camera trap locations covering $30$ classes of animals, curated from data provided by the United States Geological Survey and the National Park Service. We follow the CCT-20 data split laid out in \cite{Beery_2018_ECCV}, which was explicitly designed for in-depth generalization analysis. The split uses a subset of $57,868$ images from $20$ camera locations covering $15$ classes in CCT to simultaneously investigate performance on locations seen during training and generalization performance to new locations. Bounding-box annotations are provided for all images in CCT-20, whereas the rest of CCT has only class labels. In the CCT-20 data split, \textit{\bf cis-locations} are defined as locations seen during training and \textit{\bf trans-locations} as locations not seen during training (see Fig.\ref{fig:annotPerLoc}). Nine locations are used for trans-test data, one location for trans-validation data, and data from the remaining $10$ locations is split between odd and even days, with odd days as cis-test data and even days as training and cis-validation data (a $95\%$ of data from even days for training, $5\%$ for testing).
\begin{figure}[!ht]
   \centering
   \subfloat[Training images]{\includegraphics[width=.33\linewidth]{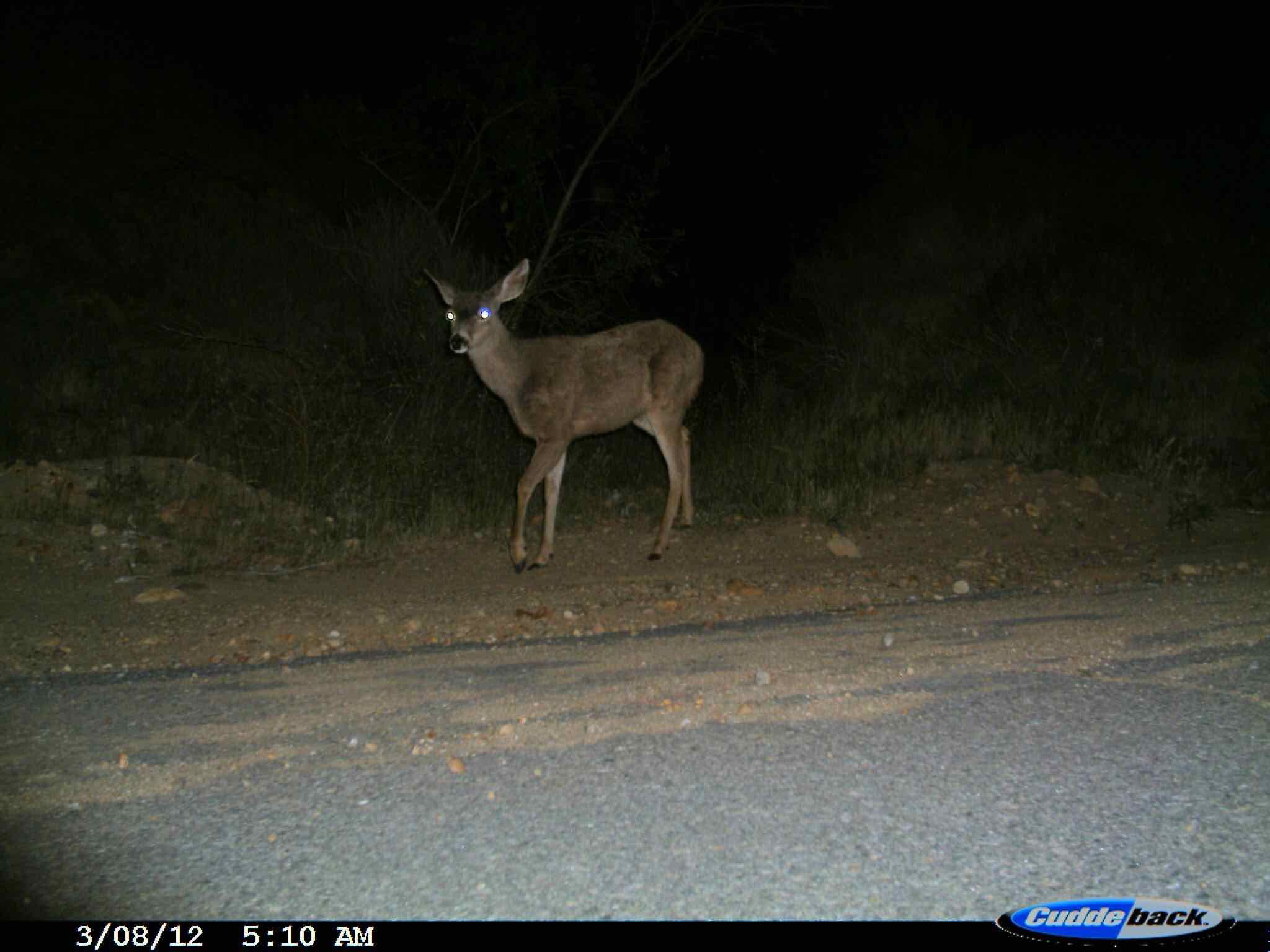}\quad \includegraphics[width=.33\linewidth]{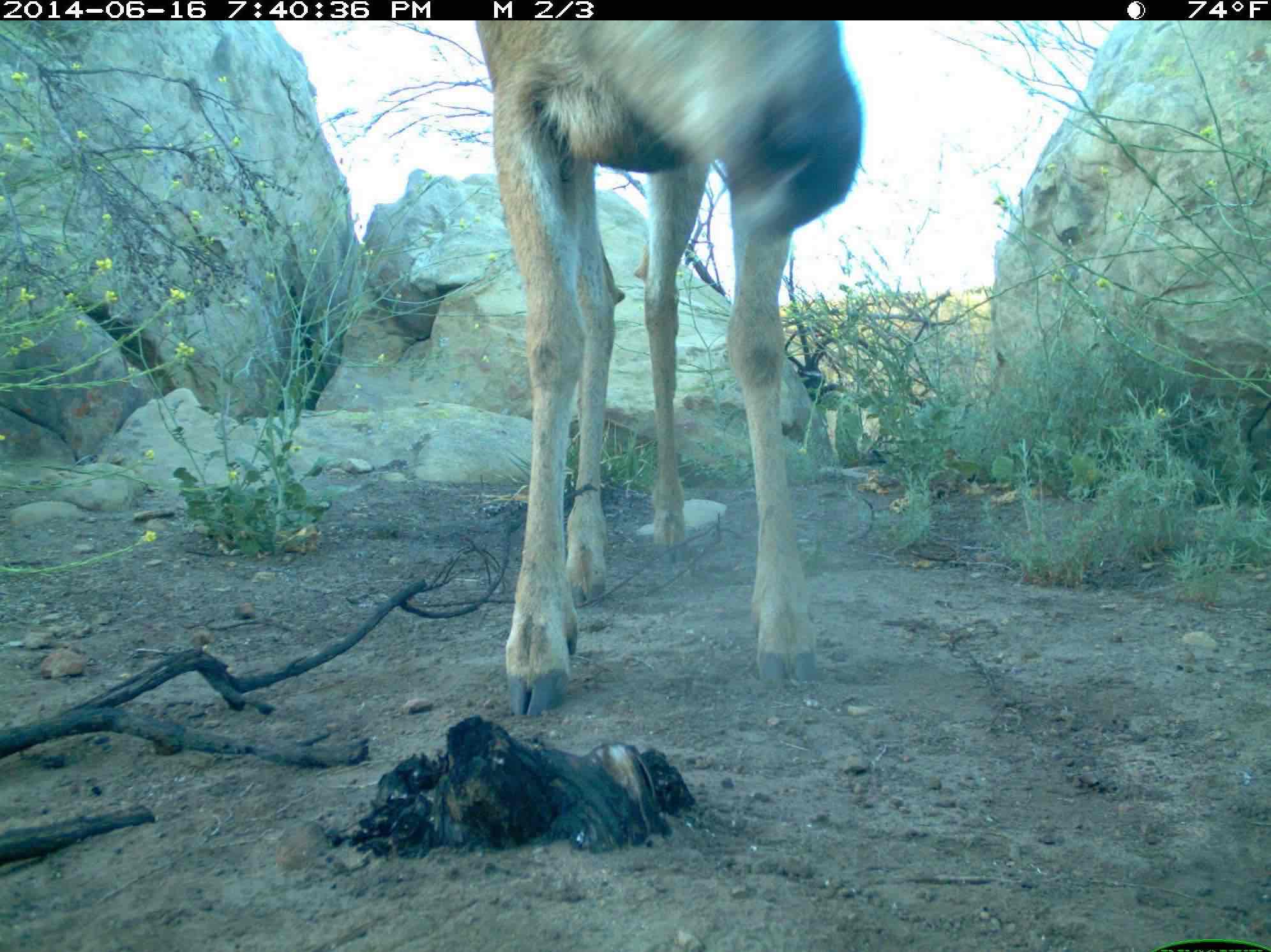}}\\
   \subfloat[Cis test images]{\includegraphics[width=.33\linewidth]{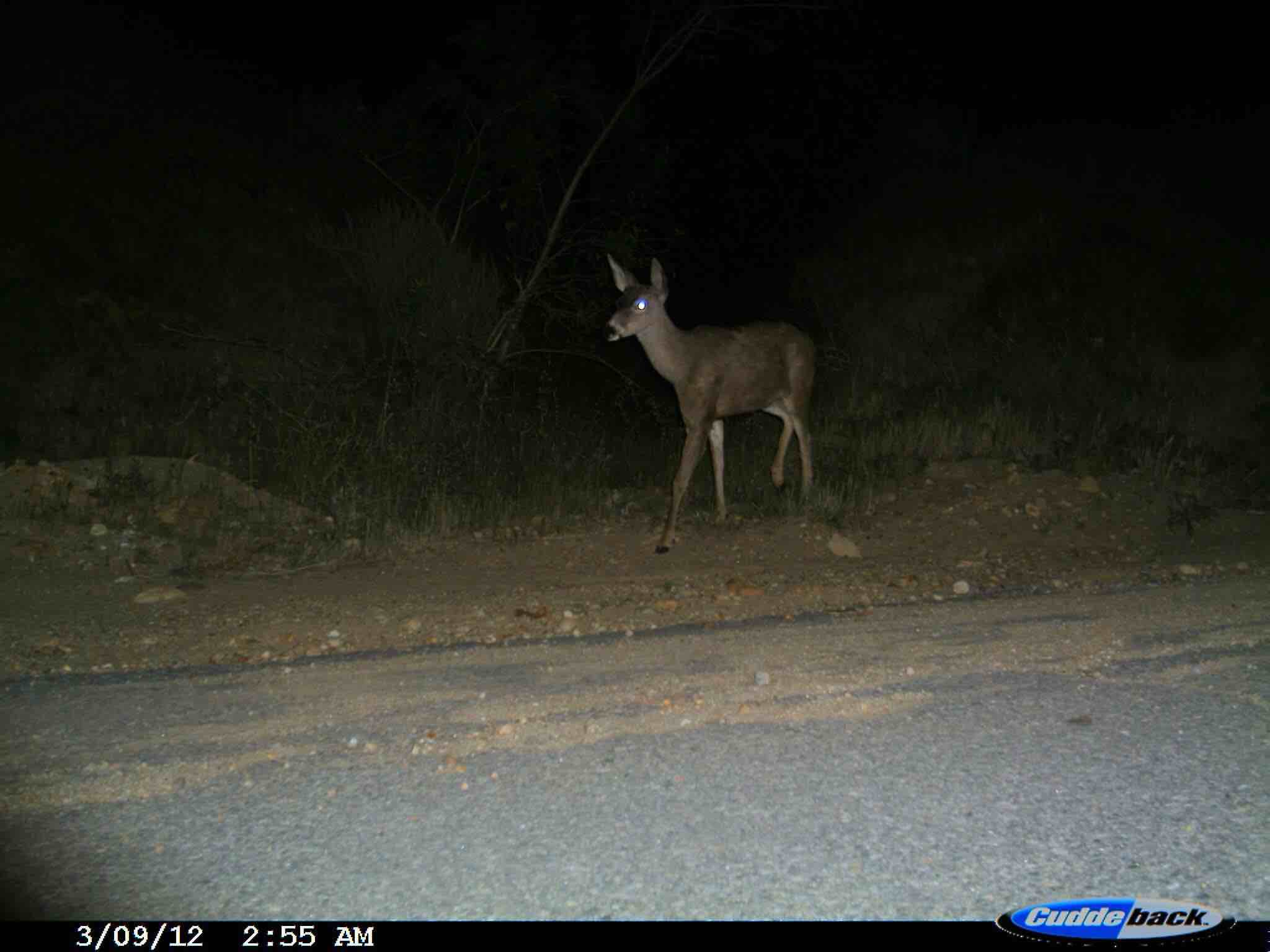}\quad \includegraphics[width=.33\linewidth]{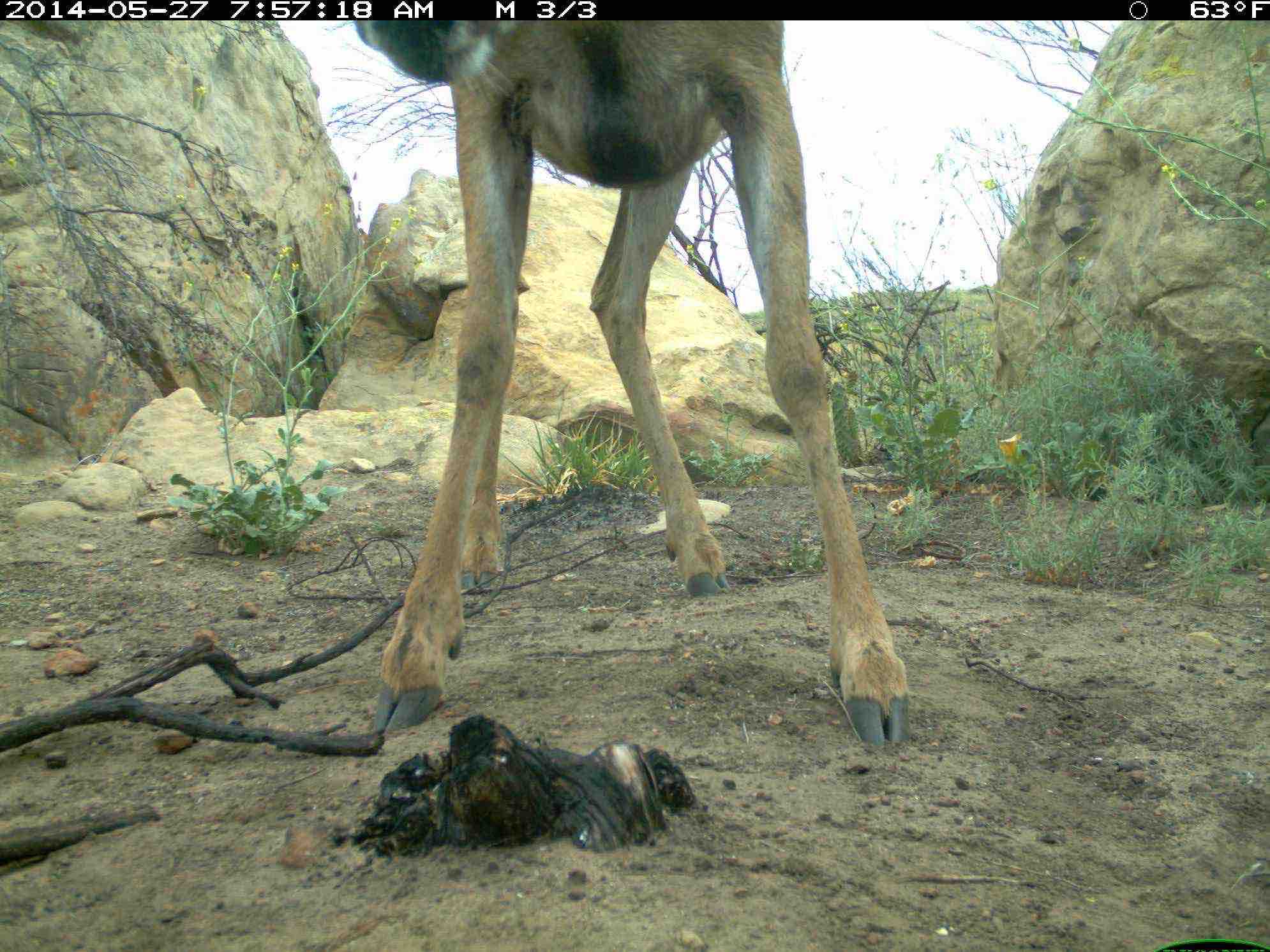}}\\
   \subfloat[Trans+ test images]{\includegraphics[width=.33\linewidth]{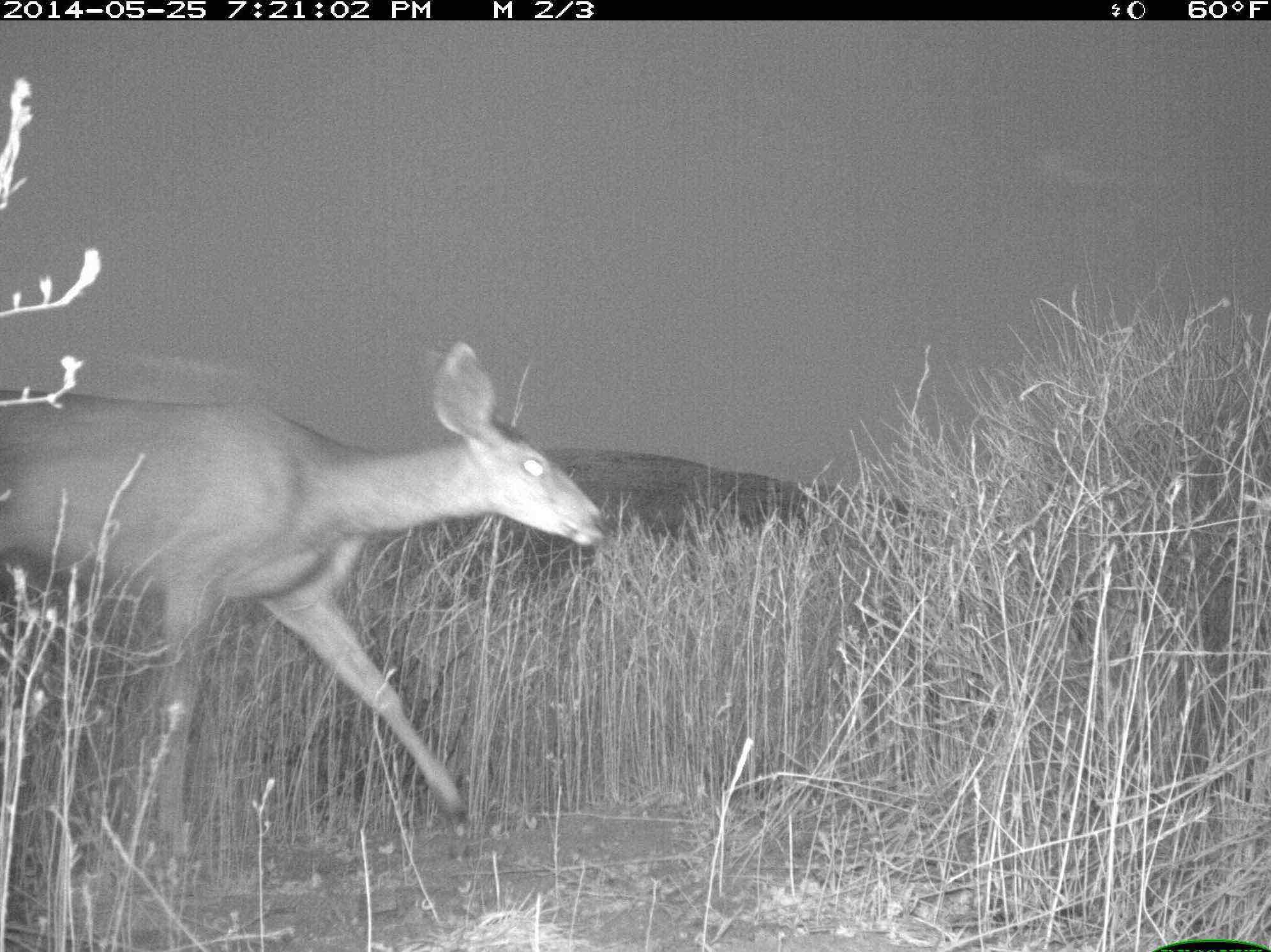}\quad \includegraphics[width=.33\linewidth]{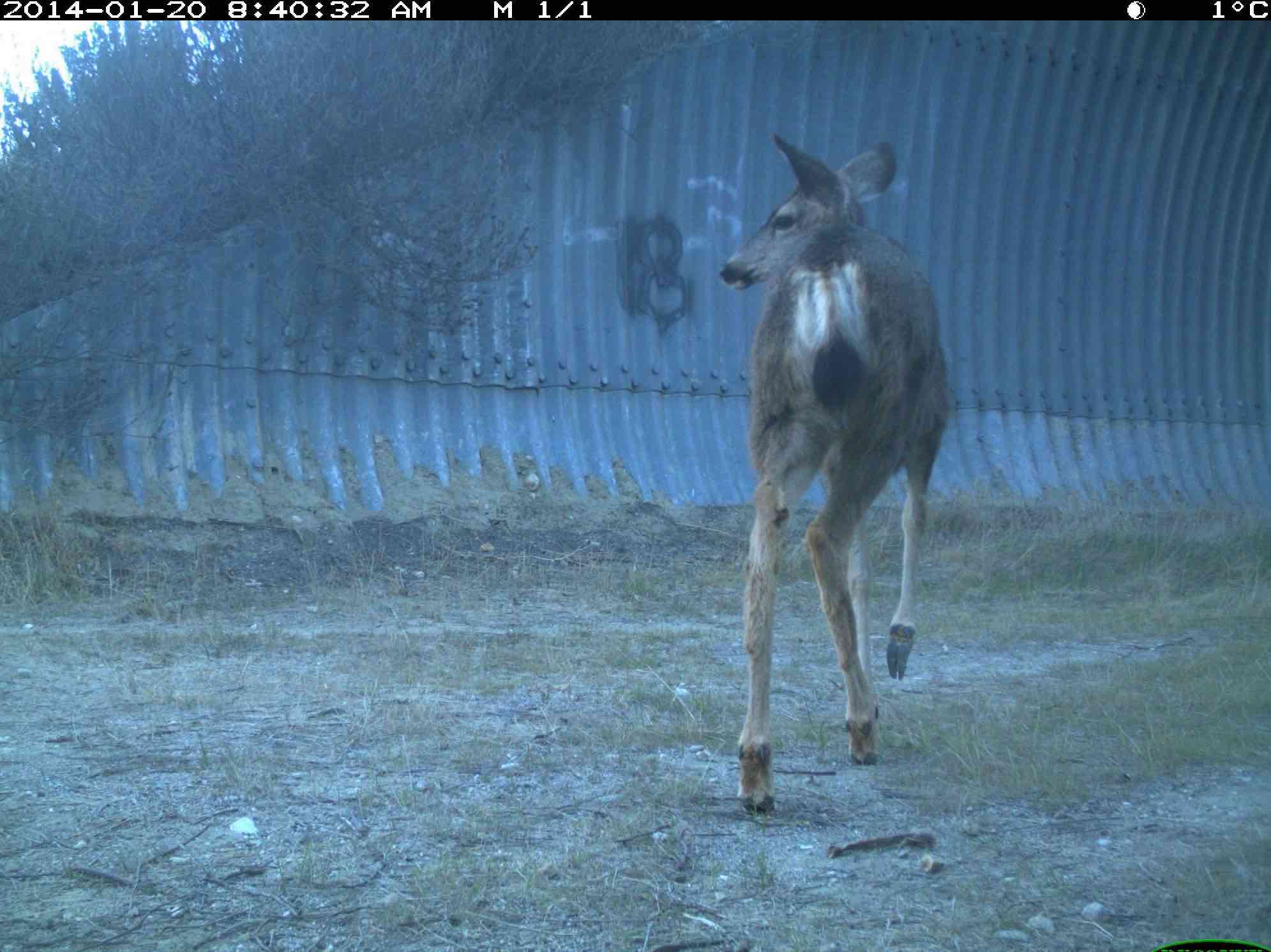}}\\
   \subfloat[iNaturalist images]{\includegraphics[width=.33\linewidth]{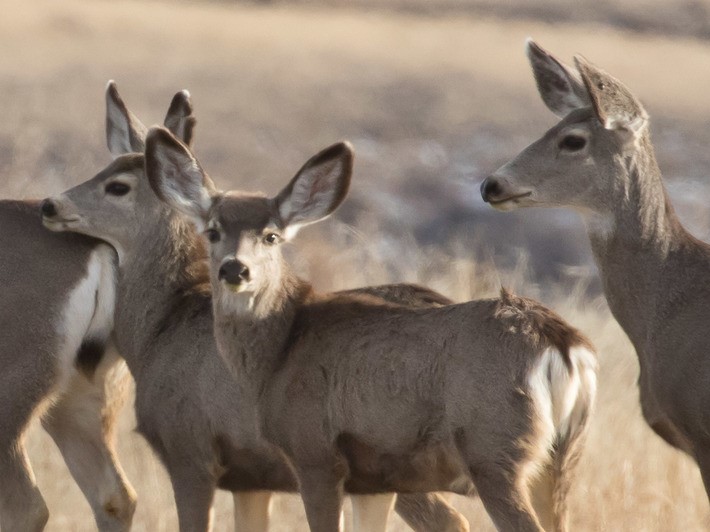}\quad \includegraphics[width=.33\linewidth]{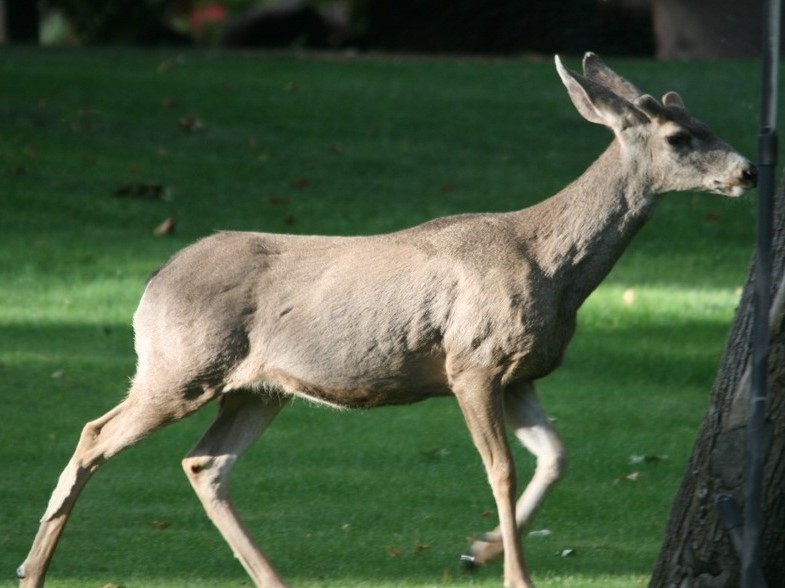}}
   \setlength{\belowcaptionskip}{-20pt}
   \caption{\textbf{Cis vs. Trans:} The cis-test data can be very similar to the training data: animals tend to behave similarly at a single location even across different days, so the images collected of each species are easy to memorize intra-location. The trans data has biases towards specific angles and lighting conditions that are different from those in the cis locations, and as such is very hard to learn from the training data. iNaturalist data represents a domain shift to human-curated images.}
   \label{fig:cis_v_trans_ims}
\end{figure}
In order to study the effect of simulated data on rare species, we focus on deer, which are rare in CCT-20, with only $44$ deer examples out of the $13,553$ images in the training set (see Fig.\ref{fig:annotPerLoc}). In order to focus on the performance of a single rare class, we remove the other two rare classes in CCT-20: badgers and foxes. We noted that there were no deer images in the established CCT-20 trans sets. In reality, deer are far from uncommon: unlike a truly rare species, there exist sufficient images of deer in the CCT dataset outside of the CCT-20 locations to rigorously evaluate performance. To facilitate deeper investigation of generalization we collected bounding-box annotations for an additional $16$K images from CCT across $65$ new locations, which we add to the trans-validation and trans-test sets to cover a wider variety of locations and classes (including deer). We call this augmented trans set \textit{trans+} (see Fig.\ref{fig:annotPerLoc}) and will release the annotations at publication. To further analyze generalization, we also test on data containing deer from the iNaturalist 2017 dataset \cite{van2017inaturalist}, which represents a domain shift to human-captured and human-selected photographs. We consider \textit{Odocoileus hemionus} (mule deer) and \textit{Odocoileus} virginianus (white-tailed deer) images from iNaturalist, the two species of deer seen in the CCT data.

\begin{figure}
\subfloat{\includegraphics[width=6.6cm]{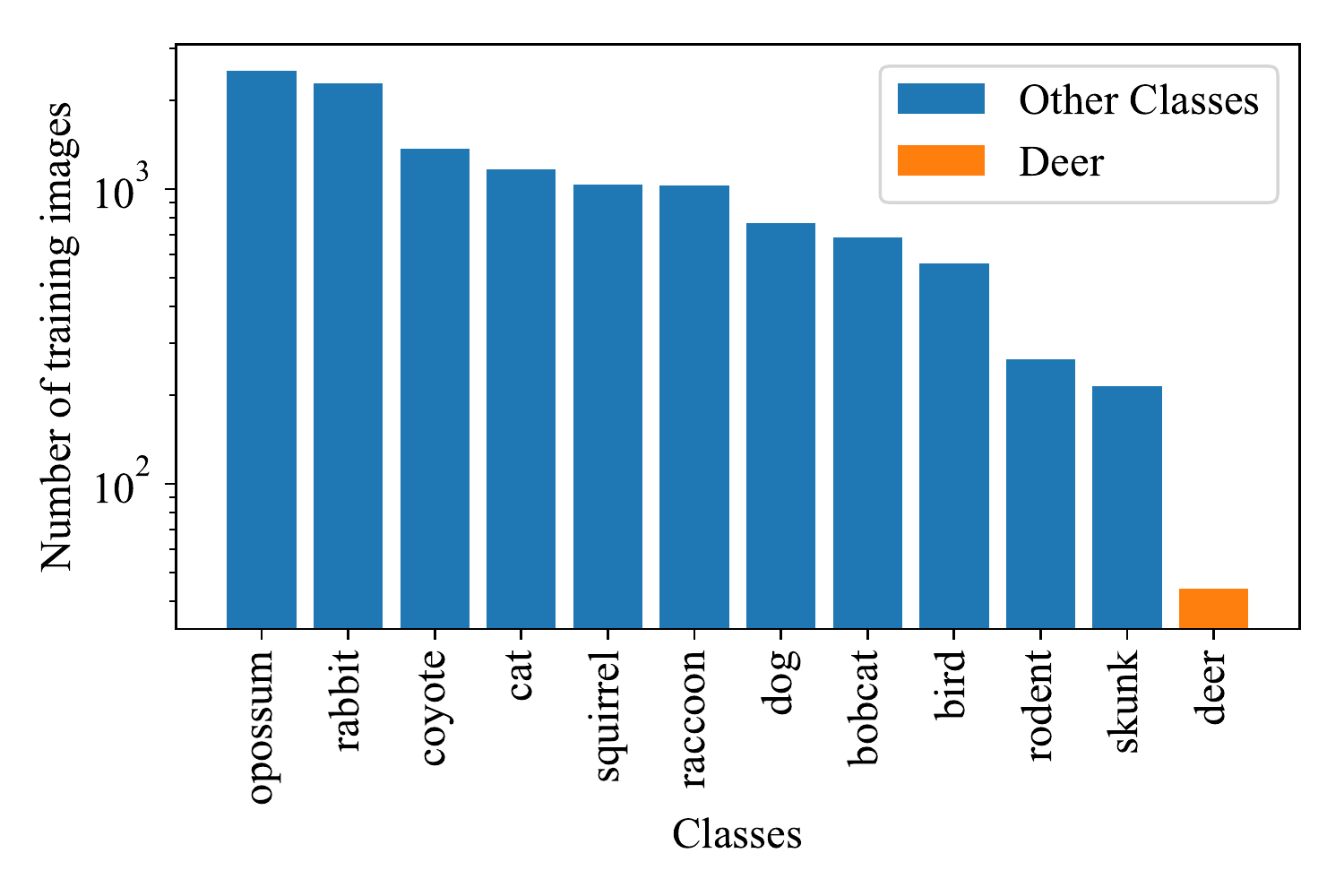}}\\
\vspace{-15pt}
\subfloat{\includegraphics[width=6.6cm]{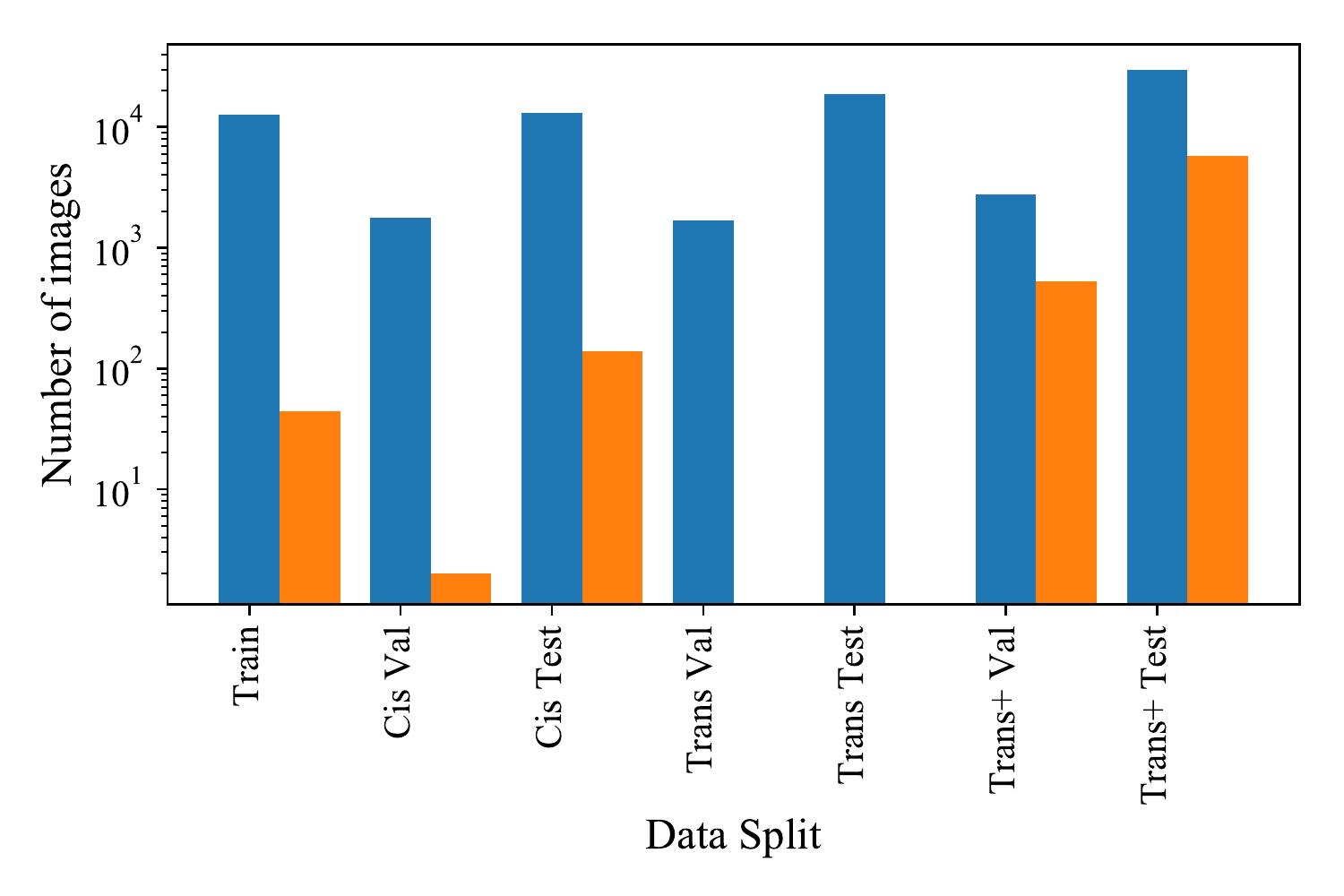}}
\vspace{-10pt}
\setlength{\belowcaptionskip}{-20pt}
\caption{\textbf{(Top) Number of training examples for each class.} Deer are rare in the training locations from the CCT-20 data split. We focus on deer as a test species in order to investigate whether we can improve performance on a ``rare" class. Since deer are not rare at other camera locations within the CCT dataset, we have enough test data to thoroughly evaluate the effect. \textbf{(Bottom) Number of examples for each data split, for deer and other classes.} In the CCT-20 data split there were no trans examples of deer. We added annotations to the trans val and test sets for an additional 16K images across 65 new locations from CCT, including 6K examples of deer. We call these augmented sets \textit{trans+}.}
\label{fig:annotPerLoc}
\end{figure} 

\Subsection{Synthetic Data}
To assess generality we leverage multiple collections of woodland and animal models to create two simulation environments, which we call TrapCam-Unity and TrapCam-AirSim. Both simulation environments and source code to generate images will be provided publicly, along with the data generated for this paper. To synthesize daytime images we varied the orientation of the simulated sun in both azimuth and elevation. To create images taken at night we used a spotlight attached to the simulated camera to simulate a white-light or IR flash and qualitatively match the low color saturation of the nighttime images. To simulate animals' eyeshine (a result of the reflection of camera flash from the back of the eye), we placed small reflective balls on top of the eyes of model animals. 

\Subsubsection{TrapCam-AirSim}
In this generation method we create a modular natural environment within Microsoft AirSim~\cite{shah2018airsim} that can be randomly populated with flora and fauna.  The distribution and types of trees, bushes, rocks, and logs can be varied and randomly seeded to create a diverse set of landscapes, from an open plain to a dense forest. We used various off-the-shelf components such as an animal pack from Epic Studios \cite{epicstudios} (Animals Vol 01: Forest Animals by GiM \cite{gim}), background terrain also from Unreal Marketplace \cite{UNREAL}, vegetation from SpeedTree \cite{speedtree}, and rocks/obstructions from Megascans \cite{megascans}. The actual area of the environment is small, at $50$ meters, but the modularity allows many possible scenes to be constructed. 

\Subsubsection{TrapCam-Unity}
In this generation method we take advantage of the ``Book of The Dead" environment \cite{bookofthedead}, a near-photorealistic, open-source forest environment published by Unity to demonstrate its high definition rendering pipeline. We move throughout the larger, fixed environment to collect data with various background scenes. We include $17$ animated deer models from five model sets, including the GiM models used in TrapCam-AirSim. 

\Subsubsection{Simulated animals on empty images}
Similar to the data generated in \cite{Varol_2017_CVPR}, we generate synthetic images of deer by rendering deer on top of real camera trap images containing no animals, which we call \textit{Sim on Empty} (see Fig.\ref{fig:sim_methods}). We first generated animal foreground images by randomizing the location, orientation in azimuth, pose and illumination of the deer, then pasted the foreground images on top of the real empty images. A limitation is that the deer are not in realistic relationships or occlusion scenarios with the environment around them. We also note that the empty images used to construct this data come from both cis and trans locations, so Sim on Empty contains information about test-set backgrounds unavailable in the purely simulated sets. This choice is based on current camera trap literature, which first detects the presence of any animal, and then determines animal species \cite{norouzzadeh2017automatically,Beery_2018_ECCV}. After the initial animal detection step, the empty images are known and can be utilized.

\Subsubsection{Segmented animals on empty images} 
We manually segmented the 44 examples of deer from the training set and pasted them at random on top of real empty camera trap images, which we call \textit{Real on Empty} (see Fig.\ref{fig:sim_methods}). This allows us to analyze whether the generalization challenge is related to memorizing the training deer+background or memorizing the training deer regardless of background. Similar to the Sim on Empty set, these images do not have realistic foreground/background relationships and the empty images come from both cis and trans locations.

\Section{Experiments}
\cite{Beery_2018_ECCV} showed that detecting and localizing the presence of an ``animal" (where all animals are grouped into a single class) both generalizes well to new locations and improves classification performance. We focus on classification of cropped ground-truth bounding boxes as opposed to training multi-class detectors in order to disambiguate classification and detection errors.  We are specifically investigating how added synthetic training data for rare classes effects model performance on both rare and common classes. 

We determined that the Inception-Resnet-V2 architecture \cite{szegedy2017inception} worked best for the cropped-box classification task by comparing performance across architectures (see Supplementary Material). Most classification systems are pretrained on Imagenet, which contains animal classes. %In order to remove the potential for bias within our class set based on pretraining (t
To ensure that our ``rare" class was truly something the model was unfamiliar with, as opposed to something it had seen in pretraining, we pretrained our classifiers on \textit{no-animal ImageNet}, a dataset we define by removing the ``animal" %, animate being, beast, brute, creature, fauna" 
 subtree (all classes under synset node n$00015388$) from ImageNet. %All experiments are trained from an Inception-Resnet-V2 model, which is pretrained on no-animal ImageNet. 
 We use an initial learning rate of $0.0045$, RMSprop with a momentum of $0.9$ \cite{tieleman2012lecture}, and a square input resolution of $299$. We employ random cropping (containing at least $65$\% of the region), horizontal flipping, color distortion, and blur as data augmentation. Model selection is performed using early stopping based on trans+ validation set performance ~\cite{bengio2012practical}.

\Subsection{Effect of increase in simulated data} 
\begin{figure}[t]
\centering
\hspace{-0.25in}\includegraphics[width=8.5cm]{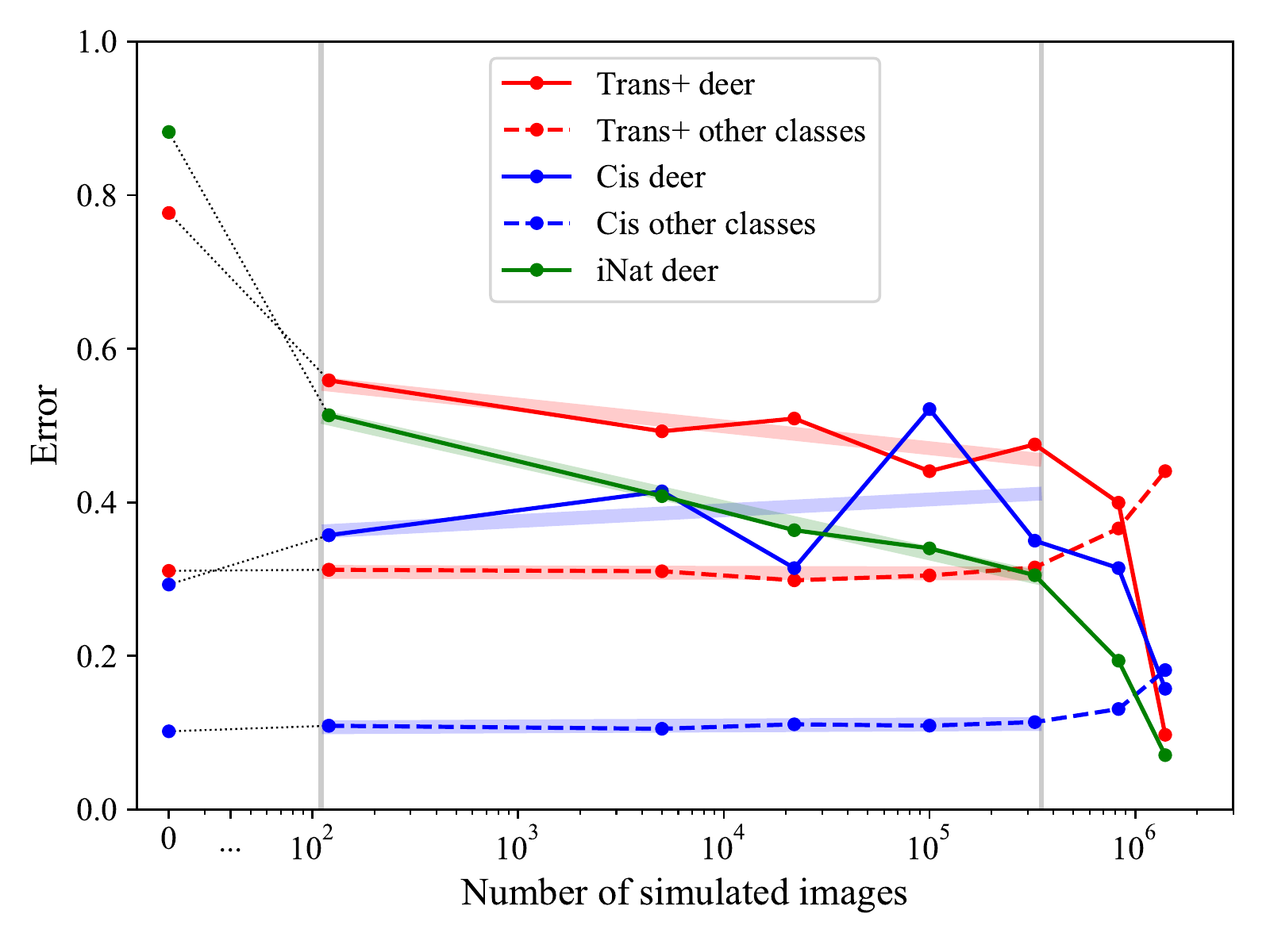}
\vspace{-10pt}
\setlength{\belowcaptionskip}{-15pt}
\caption{\textbf{Error as a function of number of simulated images seen during training. We divide this plot into three regions.} The leftmost region is the baseline performance with no simulated data, shown at x=0. Each point is connected with a grey dotted line to the corresponding experiments (due to the x-axis being in log scale). 
The middle region is an area where additional simulated training data increases performance on the rare class and does not harm the performance of the remaining classes, trend lines are visualized for this region. The rightmost region, where many simulated images are added to the training set, results in a biased classifier, hurting the performance of the other classes. We compare the class error for ``deer" and ``other classes" in both the ``cis" and ``trans+" testing regimes. Lines marked ``deer" use only the deer test images for the error computation. Lines marked ``other classes" use all the images in the other classes (excluding deer) for the error computation. Error is defined as the number of incorrectly identified images divided by the number of images.
\textbf{In the leftmost region:} note the difference in baseline performance between cis deer and trans+ deer; this is our generalization challenge. 
\textbf{In the middle region:} note that performance on ``other classes" remains constant. ``Trans+ deer" performance decreases as simulated images are added, and ``Cis deer" performance increases as simulated images are added. Cis deer images are very similar to images seen during training (see Fig. \ref{fig:cis_v_trans_ims}). As simulated data is added at training time, the model is forced to learn a more complex, varied representation of deer.
\textbf{In the rightmost region:} note the sharp decrease in deer error and an increase in ``other classes" error across both cis and trans+. We attribute this behavior to a biased classifier, see Fig.\ref{fig:pr} (b-c) for details.
}
\label{fig:increase}
\end{figure}

\begin{figure}[t]
\centering
\captionsetup{skip=1pt}
\subfloat[Trans+ deer precision-recall curves]{\includegraphics[width=7cm]{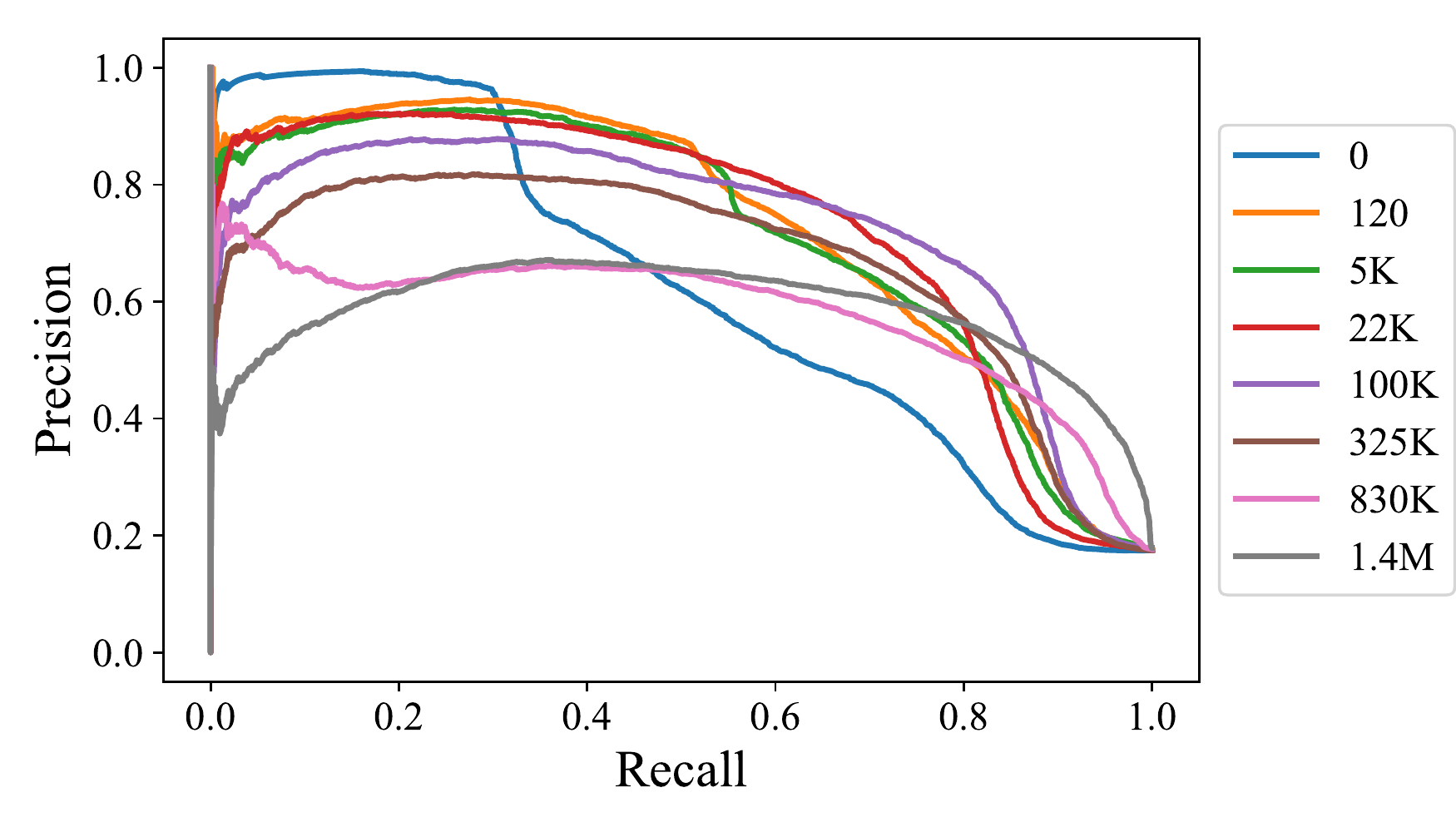}}\\
\vspace{-10pt}
\subfloat[Confusion matrix: 100K]{\includegraphics[width=4cm]{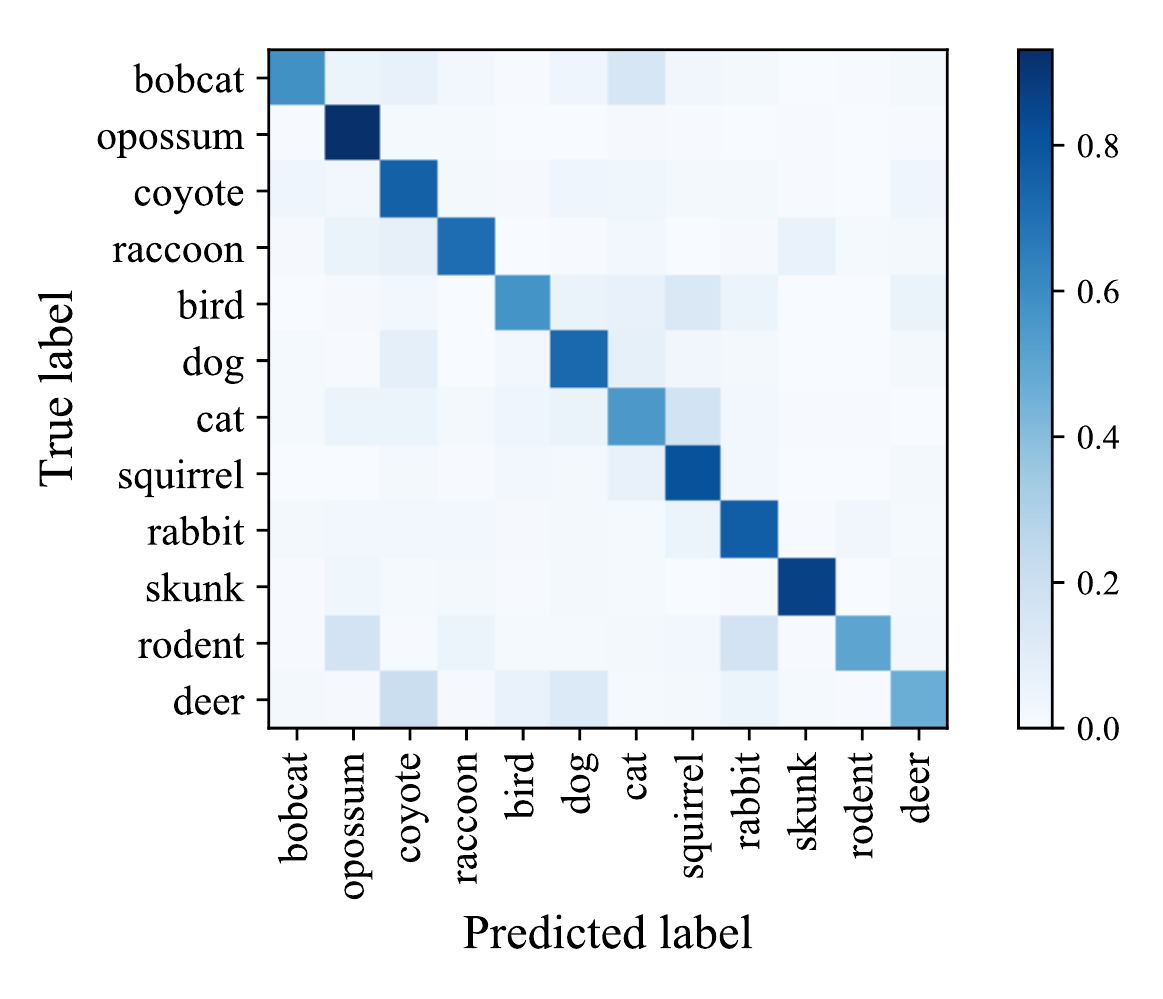}}
\subfloat[Confusion matrix: 1.4M]{\includegraphics[width=4cm]{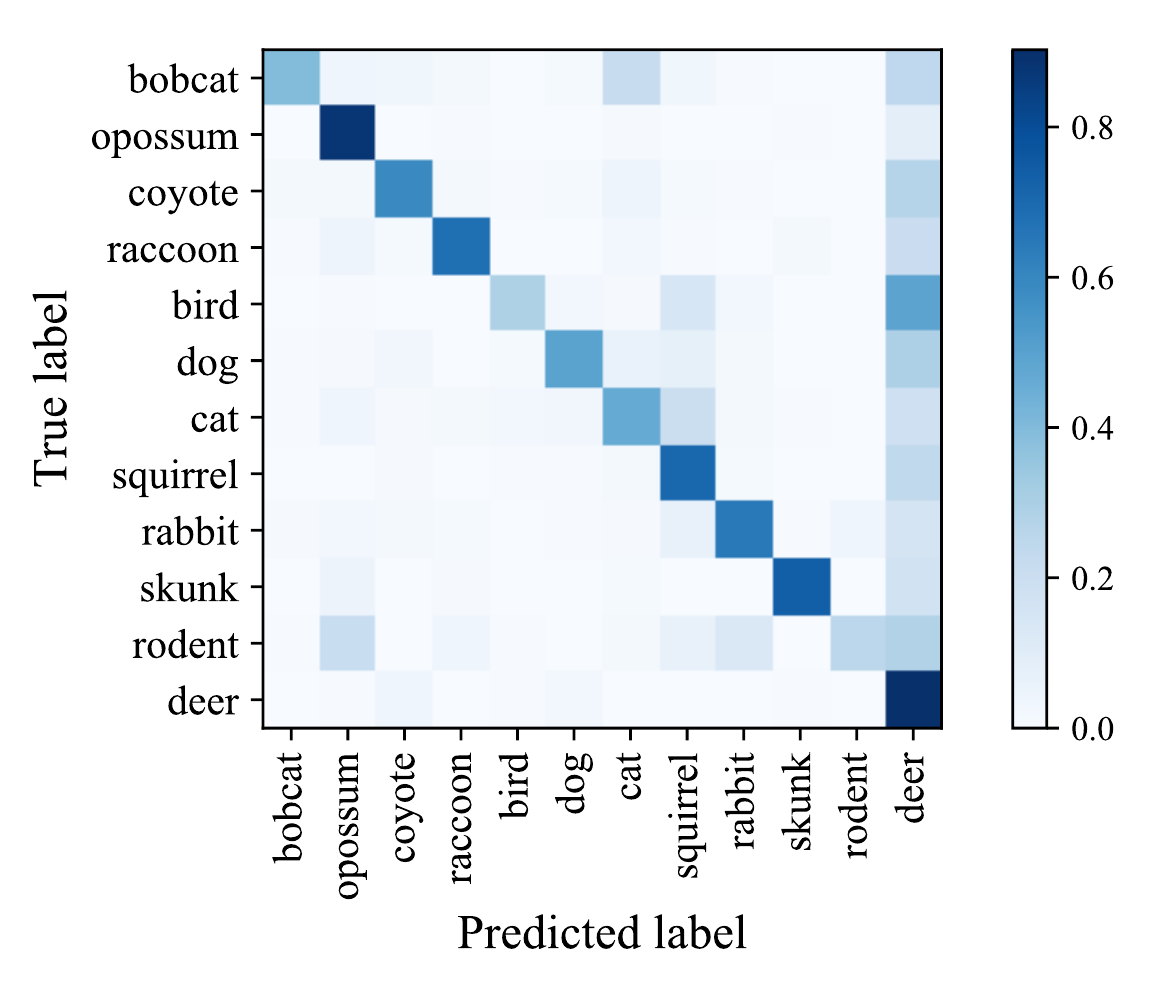}}
\vspace{-5pt}
\setlength{\belowcaptionskip}{-10pt}
\caption{\textbf{(a) Trans+ PR curves for the deer class:} Note the development of a biased classifier as we add simulated training data. The baseline model (in blue) has high precision but suffers low recall. The model trained with 1.4M simulated images (in grey) has higher recall, but suffers a loss in precision. 
\textbf{(b-c) Evidence of a biased classifier:} Compare the deer column in the confusion matrices, the model trained with 1.4M simulated images predicts more test images as deer.} 
\label{fig:pr}
\end{figure}

\begin{figure}[t]
\centering
\includegraphics[width=8.5cm]{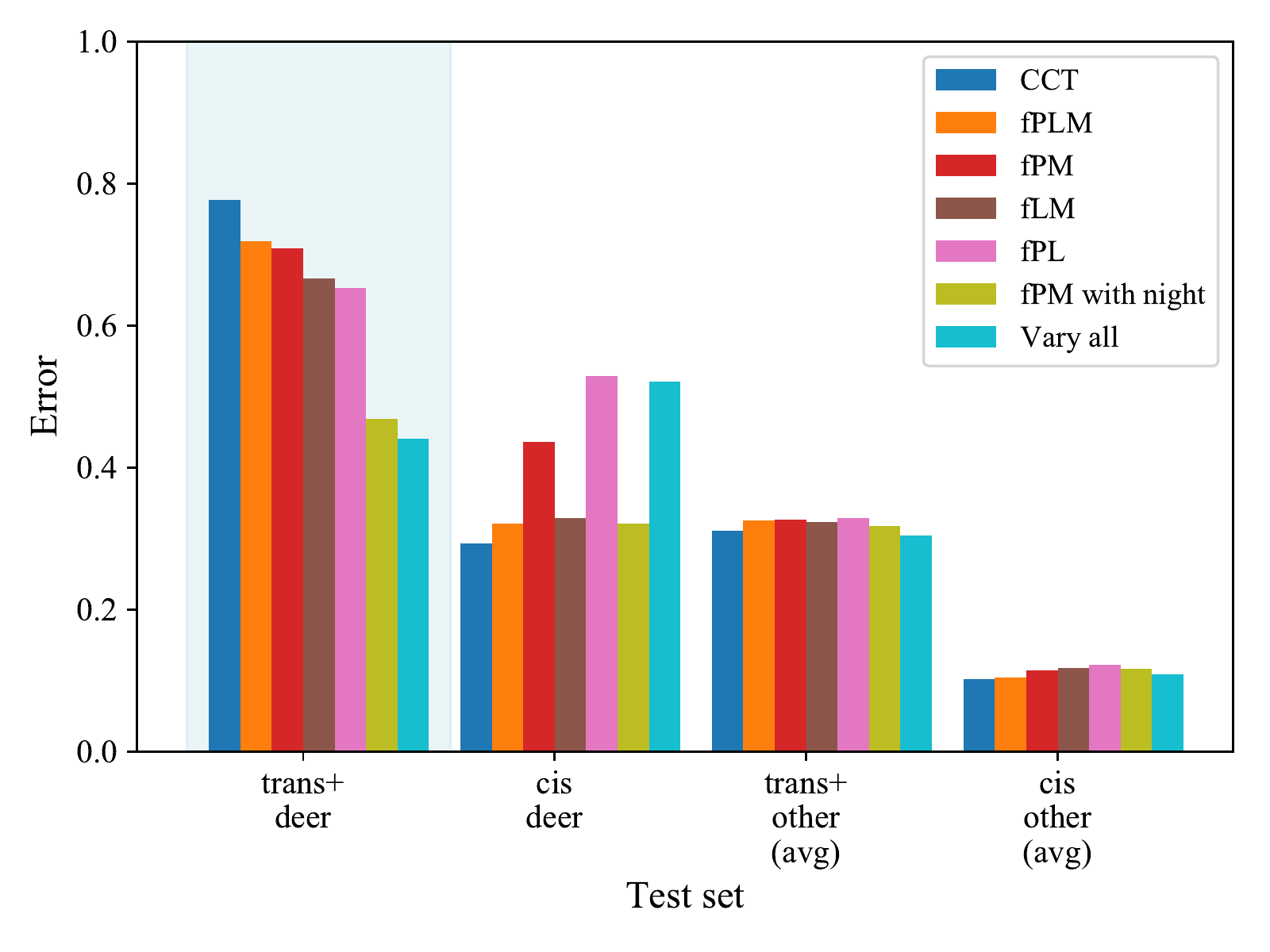}
\vspace{-15pt}
\setlength{\belowcaptionskip}{-12pt}
\caption{{\bf Error as a function of variability of simulated images seen during training: 100K simulated deer images.} Error is calculated as in Fig.\ref{fig:increase}. Trans+ deer performance is highlighted. In the legend ``CCT" means the model was trained only on the CCT-20 training set with no added simulated data. ``P" means ``pose," ``L" means ``lighting," and ``M" means ``model," while the prefix ``f" for ``fixed" denotes which of these variables were controlled for a particular experiment. For example ``fPM" means the pose and the animal model were held fixed, while the lighting was allowed to vary. The variability of simulated data is extremely important, and that while all axes of variability matter, simulating nighttime images has the largest effect.}
\label{fig:variation}
\end{figure}

\begin{figure}[t]
\centering
\includegraphics[width=8.5cm]{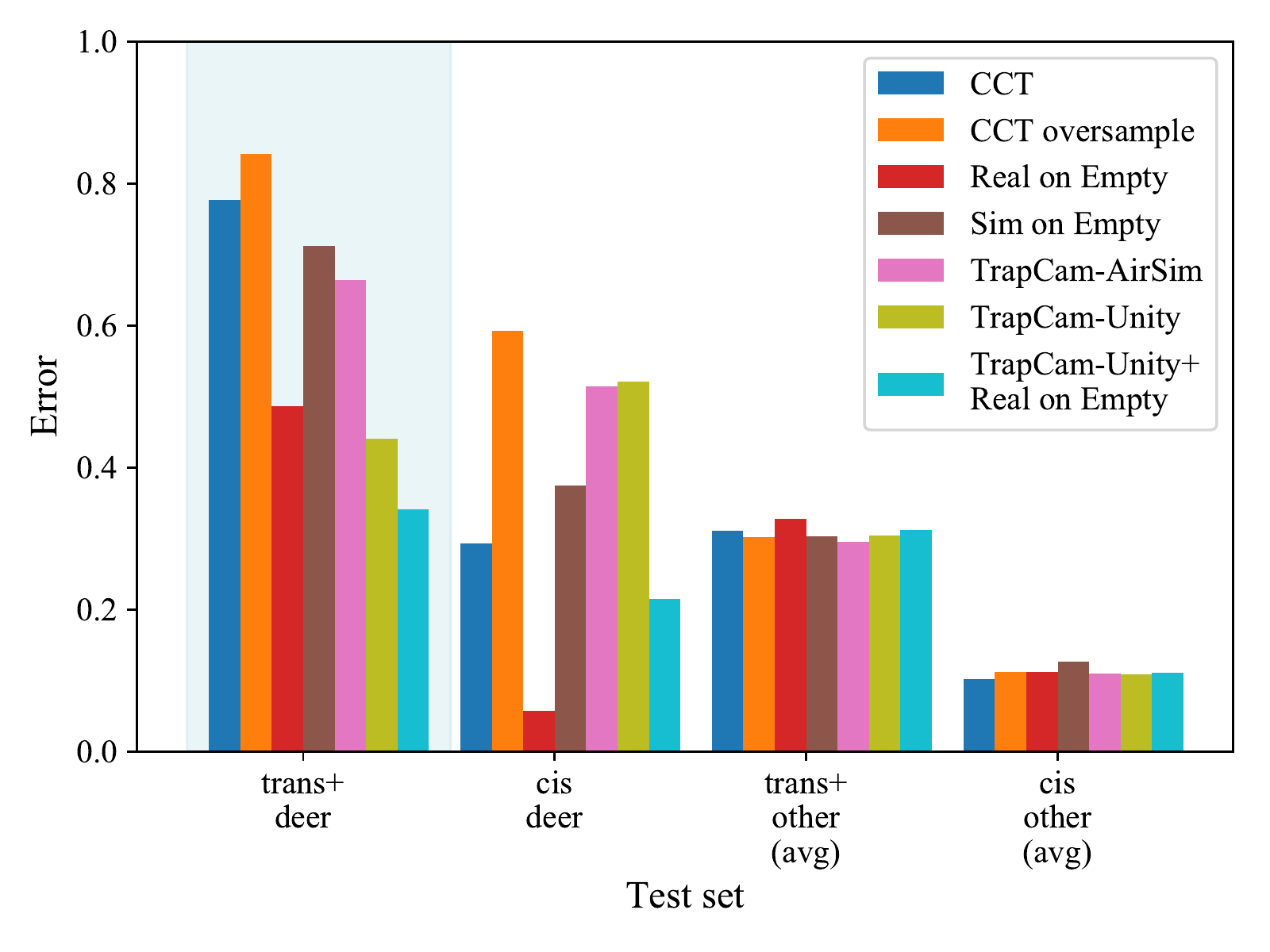}
\vspace{-15pt}
\setlength{\belowcaptionskip}{-12pt}
\caption{{\bf Error as a function of simulated data generation method: 100K simulated deer images.} Per-class error is calculated as in Fig.\ref{fig:increase}. Trans+ deer performance is highlighted. Oversampling decreases performance, and there is a large boost in performance from incorporating real segmented animals on different backgrounds (Real on Empty). TrapCam-Unity with everything allowed to vary (model, lighting, pose, including nighttime simulation) gives us slightly better trans+ performance, without requiring additional segmentation annotations. Combining Real on Empty with TrapCam-Unity ($50$K of each) gives us the best trans+ deer performance.}
\label{fig:image_generator}
\end{figure}

\begin{figure*}
\begin{center}
 \subfloat[No simulated deer]{\includegraphics[width=4.9cm]{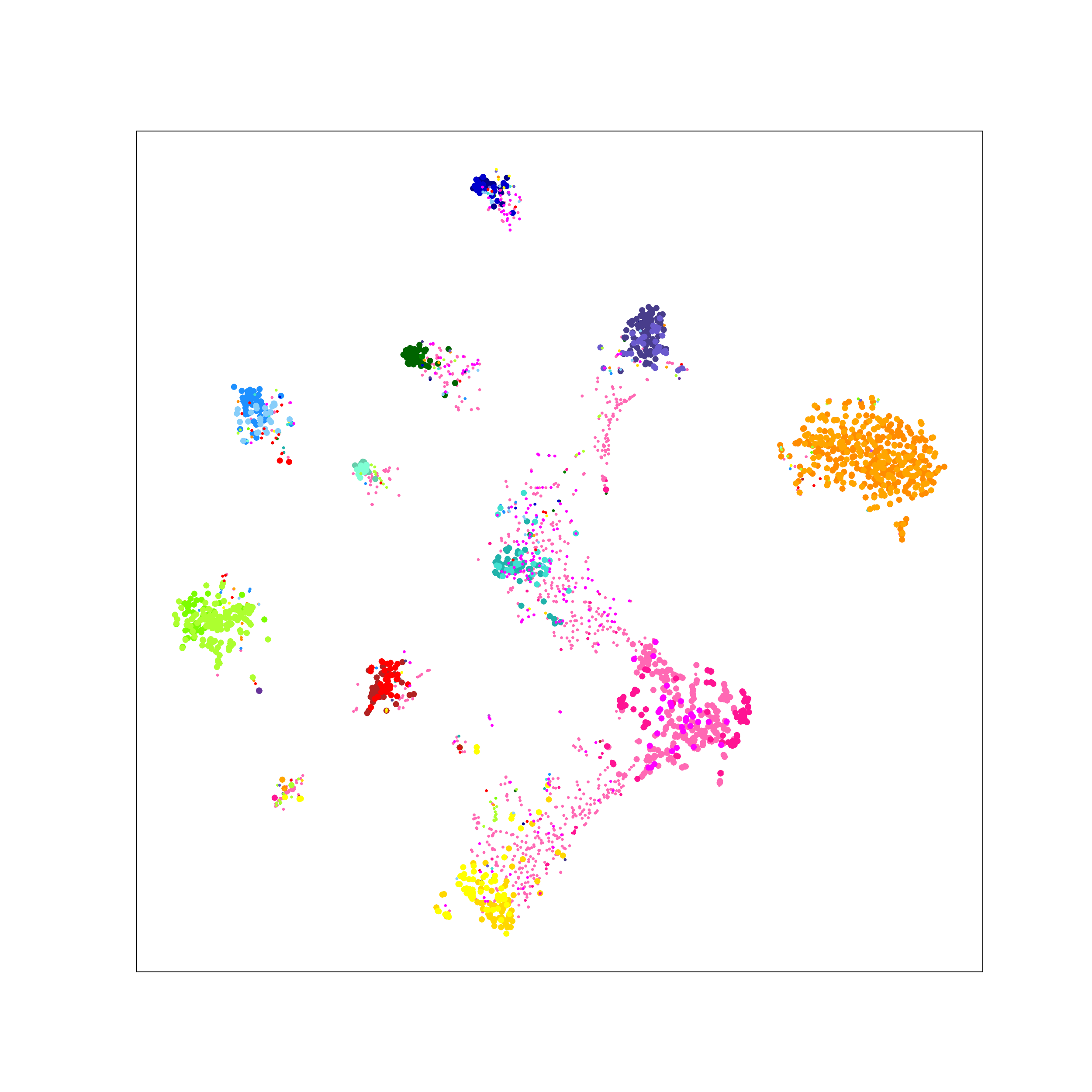}}\quad
   \subfloat[1.4M simulated deer]{\includegraphics[width=4.9cm]{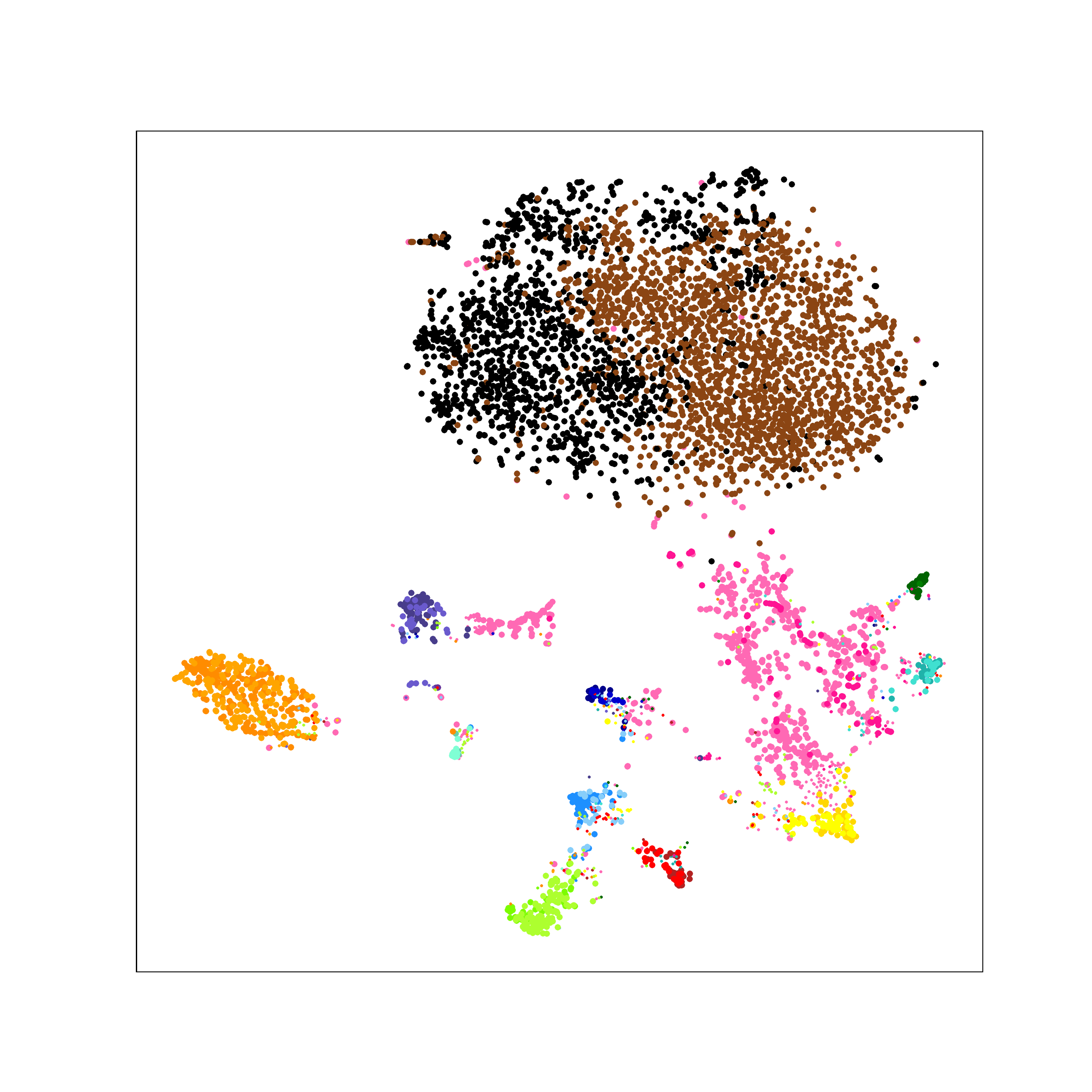}}\quad
\includegraphics[width=4cm]{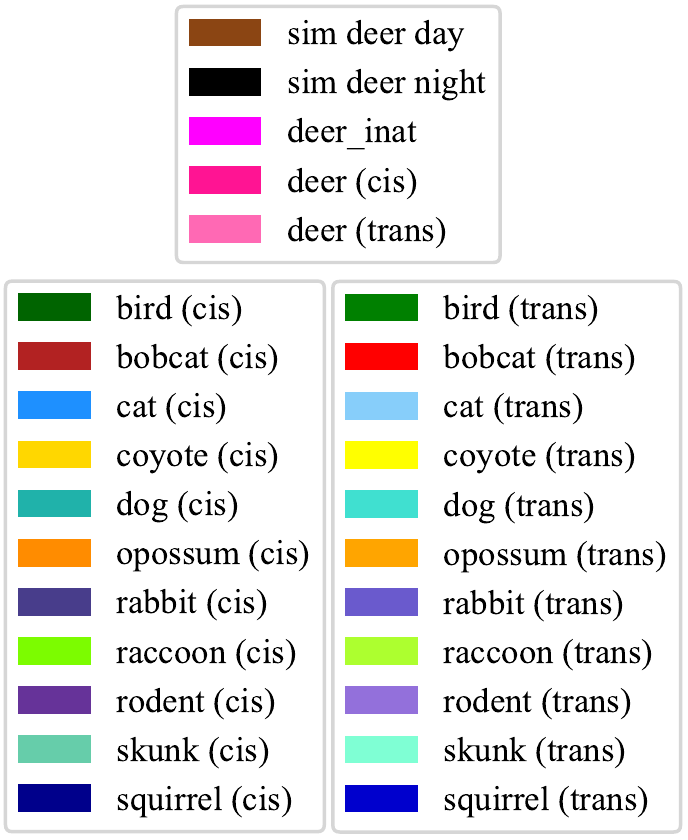}
\end{center}
\vspace{-12pt}
\setlength{\belowcaptionskip}{-15pt}
   \caption{\textbf{Visualization of network activations: more deer are classified correctly as we add synthetic data, despite the synthetic data being clustered separately.} The pink points are real deer, the brown are simulated day images and the grey are simulated night images. Large markers are points that are classified correctly, while small markers are points classified incorrectly. The plots were generated by running $200$-dimensional PCA over the activations at the last pre-logit layer of the network when running inference on the test sets, and then running 2-dimensional tSNE over the resulting PCA embedding.}
\label{fig:visualization}
\end{figure*}

We explore the trade-off in performance when increasing the number of simulated images, from $120$ to $1.4$ million (see Fig.\ref{fig:increase}). Very little simulated data is needed to see a trans+ performance boost: with as few as $120$ simulated images we see a $28$\% decrease in per-class error on trans+ deer, with $<0.5$\% increase in average per-class error on the other trans+ classes. As we increase the number of simulated images, trans+ performance improves: with $100$K simulated images we see a $39$\% decrease in trans+ deer error, with $<0.5$\% increase in error for the other trans classes. There exists some threshold ($>325$K) where, if passed, an increase in simulated data noticeably biases the classifier towards the deer class (see Fig.\ref{fig:pr}): with $1.4$ million simulated images, our trans+ deer error decreases by $88$\%, but it comes at the cost of a $13$\% increase in average per-class error across the other classes. At this point there is an overwhelming class prior towards deer: the next-largest class at training time would be opossums with $2,514$ images, $3$ orders of magnitude less. 

Surprisingly, cis deer performance decreases with added simulated data. Although the images were taken on different days (train from even days, cis-test from odd days) the animals captured were to some extent creatures of habit. This results in training and test images that are nearly identical from within the same locations (see Fig.\ref{fig:cis_v_trans_ims}). Almost all cis test deer images have at least one visually similar training image. As simulated data is added at training time, the model is forced to learn a more complex, varied representation of deer. As a result, we see cis deer performance decrease. To quantify robustness, we ran the $100$K experiment three times. We found that trans+ deer error had a standard deviation of $2\%$ and cis deer error had a standard deviation of $4\%$, whereas the average error across other classes had a standard deviation of $0.2\%$ for both cis and trans. 

We also investigate performance on deer images from iNaturalist \cite{van2017inaturalist}, which are individually collected by humans and are usually relatively centered and well-focused (and therefore easier to classify) but represent a domain shift (see Fig.\ref{fig:cis_v_trans_ims}). Adding simulated data improves performance on the iNaturalist deer images (see Fig.\ref{fig:increase}), demonstrating the robustness and generality of the representation learned.

\Subsection{Effect of variation in simulation}
In order to understand which aspects of the simulated data are most beneficial, we considered three dimensions of variation during simulation: pose, lighting, and animal model. Using the TrapCam-Unity simulator, 100K daytime simulated images were generated for each of these experiments. As a control, we created a set of data where the pose, lighting, and animal model were all fixed. We then created sets with varied pose, varied lighting, and varied animal model, each with the other variables held fixed.  An additional set of data was generated varying all of the above. Unsurprisingly, widest variation results in the best trans+ deer performance. The individual axes of variation do have an effect of performance, and some are more ``valuable" than others (see Fig.\ref{fig:variation}). There are many more dimensions of variation that could be explored, such as simulated motion blur or variation in camera perspective.  For CCT data, we find adding simulated nighttime images has the largest effect on performance. We have determined that for deer $49\%$ of training images, $53\%$ of cis test images, and $56\%$ of trans+ test images were captured at night, using either IR or white flash. Simulating only daytime images injects a prior towards deer being seen during the day.  By training on half day and half night images we match the day/night prior for deer in the data. Not all species occur equally during the day or night, some are strictly nocturnal. Our results suggest that a good strategy is to determine the appropriate ratio of day to night images using your training set and match that ratio when adding simulated data. 

\Subsection{Comparing simulated data generation methods}
We compared the performance gain from 4 methods of data synthesis, using $100$K added deer images for each (see Fig.\ref{fig:image_generator}.  The animal model is controlled (each simulated set uses the same GiM deer model for these experiments) for a fair comparison of the efficacy of each generation method. As a control, we consider oversampling of the rare class. This creates the same sampling prior towards deer without introducing any new information. Oversampling performs worse than just training on the unbalanced training set by causing the model to overfit the deer class to the training images. By manually segmenting out the deer in the $44$ training images and randomly pasting them onto empty backgrounds we see a large improvement in performance. Cis error goes down to $6$\% with this method of data augmentation, which makes sense in the view of the strong similarities between the training and cis-test data (see Fig.\ref{fig:cis_v_trans_ims}).

Real on Empty and Sim on Empty are able to approximate both ``day" and ``night" imagery, a deer pasted onto a nighttime empty image is actually a reasonable approximation of an animal illuminated by a flash at night (see Fig.\ref{fig:night}). They also have the additional benefit of using backgrounds from both cis and trans sets, giving them trans information not provided by the simulated datasets. TrapCam-Unity with all variability enabled is our best-performing model without requiring additional segmentation annotations. If segmentation information is available, Real on Empty combined with TrapCam-Unity ($50$K of each) improves both cis and trans deer performance: trans deer error decreases to $36\%$ (a $54$\% decrease compared to CCT only), with $<2\%$ increase in error on trans other classes.

\Subsection{Visualizing the representation of data}
In order to visualize how the network represents simulated data vs. real data, we used PCA and tSNE \cite{maaten2008visualizing} to cluster the activations of the final pre-logit layer of the network.  These visualizations can be seen in Fig.\ref{fig:visualization}. Interestingly, the model learns ``deer" bimodally: simulated deer are clustered almost entirely separately from real deer, with a few datapoints of each ending up in the opposite cluster.  Even though those clusters overlap only slightly, the network is surprisingly able to classify more deer images correctly.

\Section{Conclusions and Future Work}
We explored using simulated data to augment rare classes during training. Towards this goal, we developed and compared multiple sources of natural-world data simulation, explicitly measured generalization via the cis-vs-trans paradigm, examined trade-offs in performance as the number of simulated images seen during training is increased, and analyzed the effect of controlling for different axes of variation and data generation methods.

From our experiments we draw three main lessons. First: using synthetic data can considerably reduce error rates for classes that are rare, and with segmentation annotations we can reduce error rates even further by additionally randomly pasting segmented images of rare classes on empty background images. Second: as the amount of simulated data is increased, accuracy on the target class improves. However, with $1000$x more simulated data than the common classes, we see negative effects on the performance of other classes due to the high class imbalance. Third: the variation of simulated data generated is very important, and maximum variation provides maximum performance gain. 

While an increase in simulated data corresponds to an increase in target class performance, the representation of simulated data overlaps only rarely with the real data (see Fig.\ref{fig:visualization}). It remains to be studied whether embedding techniques \cite{schroff2015facenet}, domain adaptation techniques  \cite{ganin2015unsupervised,zou2018unsupervised}, or style transfer \cite{goodfellow2014generative,Shrivastava_2017_CVPR} could be used to encourage a higher overlap in representation between the synthetic and real data, and whether that overlap would in fact lead to an increase in categorization accuracy. Additionally, the bias induced by adding large amounts of simulated data could be addressed with algorithmic solutions such as those in \cite{cui2019class,elkan2001foundations,he2008learning,he2008adasyn}. We did not discuss the drawback related to model training when using large quantities of synthetic data (increased epoch time, data storage, etc.). Another line of future work could explore the merger of a dataset simulator and the classifier so that highly variable synthetic data could be requested “online” without storing the raw frames.

Simulation is a fast, interpretable and controllable method of data generation that is easy to use and easy to adapt to new classes. This allows for an integrated and evolving training pipeline: simulated data can be generated iteratively based on needs or gaps in performance. Our analysis suggests a general methodology when using simulated data to improve rare-class performance: 1) generate small, variable sets of simulated data (even small sets can drive improvement), 2) add these sets to training and analyze performance to determine ideal ratios and dimensions of variation, 3) take advantage of ease and speed of generation to create an abundance of data based on this ideal distribution, and determine an operating point of number of added simulated images to optimize performance between rare target class and other classes based on the project goal.

Further, the performance gains we have demonstrated, along with the data generation tools we contribute to the community, will allow biodiversity researchers focused endangered species to improve classification performance on their target species. 
Adding each new species to the simulation tools currently requires the assistance of a graphics artist. However, automated 3D modeling techniques, such as those proposed in \cite{kanazawa2018learning,reinert2016animated,cashman2013shape,pahde2019low}, might eventually become an inexpensive and practical source of data to improve few-shot learning.

The improvement we have found in rare-class categorization is encouraging, and the release of our data generation tools and the data we have generated will provide a good starting point for other researchers studying imbalanced data, simulated data augmentation, or natural-world domains.
\section{Acknowledgements}
We would like to thank the USGS and NPS for providing data. This work was supported by NSFGRFP Grant No. 1745301, the views are those of the authors and do not necessarily reflect the views of the NSF. Compute provided by Microsoft AI for Earth and AWS.

{\small
\bibliographystyle{ieee}
\bibliography{main}
}

\newpage
\begin{center}
\huge{\textbf{Supplementary Material}}
\end{center}

\section{Architecture Selection}
To select a single classification architecture to use across our experiments, we trained three classifiers: ResNet-101 V2, Inception V3, and Inception-ResNet V2. All three classifiers were pretrained on \textit{no-animal ImageNet} then trained on the Caltech Camera Traps (CCT) training set (described in the main paper, Section 3.1) with no added simulated images.  We found that Inception-ResNet V2 performed best on deer in cis and trans scenarios (see Table \ref{table:architecture}), so we decided to use Inception-ResNet V2 as the base architecture for all further experiments. 

\setlength{\tabcolsep}{4pt}
\begin{table}[h!]
\begin{center}
\caption{Error for different architectures. Error is defined as the number of incorrectly identified images divided by the number of images for each test set, where ``Deer" contains only deer images and ``Other" contains all non-deer images.}
\label{table:architecture}
\begin{tabular}{l|cc|cc|}
\cline{2-5}
&  \multicolumn{2}{|c|}{Cis Test} & \multicolumn{2}{|c|}{Trans+ Test} \\
\hline
\multicolumn{1}{|c|}{Architecture}& Deer & Other & Deer & Other \\
\hline
\multicolumn{1}{|c|}{Resnet 101 V2} & 47.86 & 11.18 & 88.63 & \textbf{29.76}    \\
\multicolumn{1}{|c|}{Inception V3} & 50.00 & 11.74 & 81.73 & 32.74 \\
\multicolumn{1}{|c|}{Inception Resnet V2}  & \textbf{29.28} & \textbf{10.17} & \textbf{77.69} & 31.07 \\
\hline
\end{tabular}
\end{center}
\end{table}

\section{Additional analysis}
\subsection{Analyzing the value of real images}
We find that our simulated data is sufficient to learn to recognize some deer even without real examples, though the real examples give a large boost in performance. The performance breakdown can be seen in Table \ref{table:no_real_deer}. These results are promising for both researchers studying zero-shot learning and biologists studying highly endangered species: it is possible to learn a species with no real training data. This avenue remains open for further study.

\subsection{Comparing night and day performance}
\begin{figure}[h!]
\centering
\includegraphics[width=8.5cm]{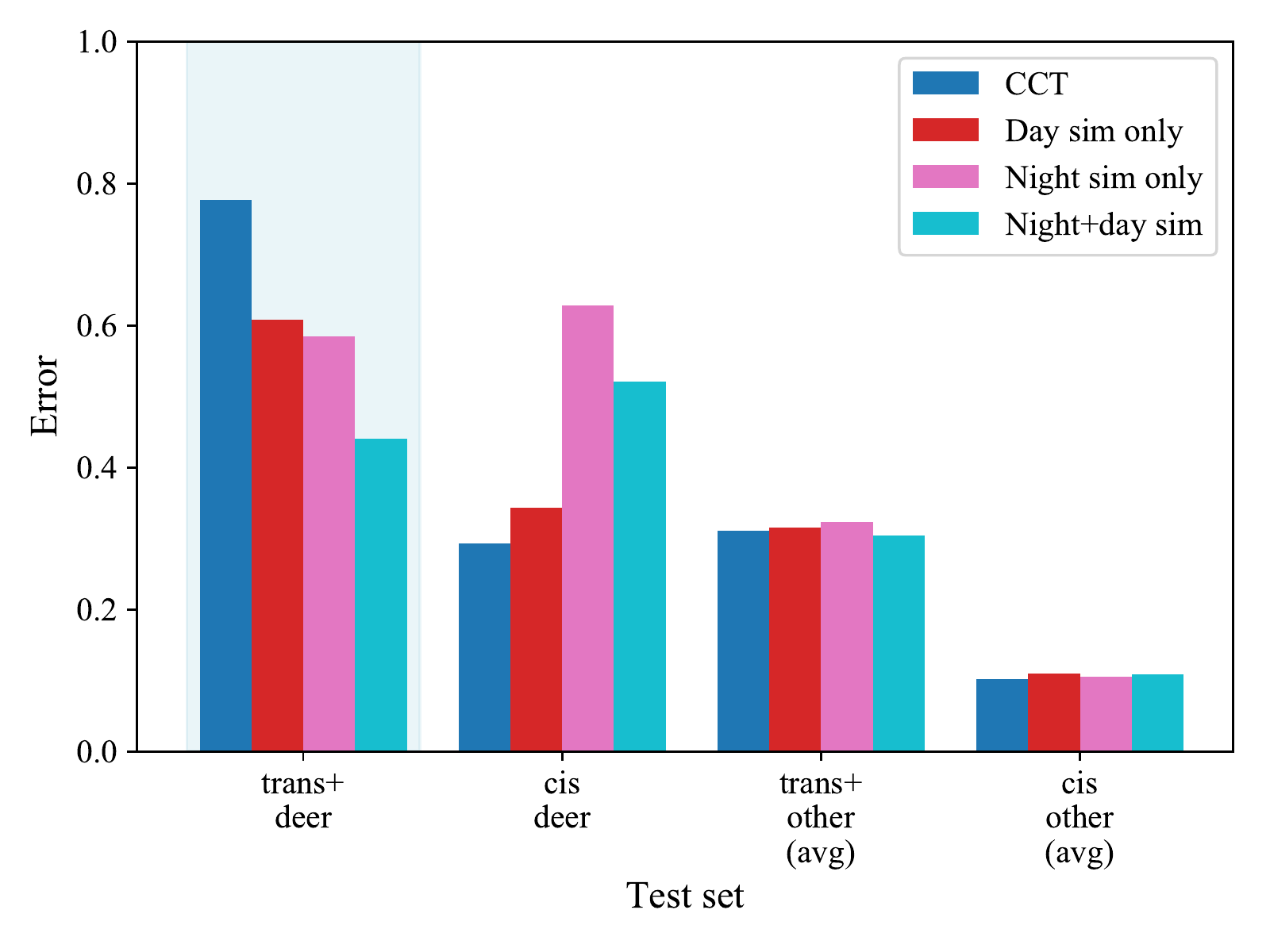}
\vspace{-15pt}
\setlength{\belowcaptionskip}{-12pt}
\caption{{\bf Error as a function of day or night simulated images: 100K simulated deer images.} Error is calculated as in Fig. 4 in the main paper. Trans+ deer performance is highlighted. Models trained on added night- or day-only simulated data perform better on trans deer than CCT alone, but the best trans deer performance comes from the 50/50 day/night split of added simulated data.}
\label{fig:day_v_night}
\end{figure}

We further analyze the effect of day and night simulation by comparing three experiments: one trained with only simulated daytime images, one trained with only simulated nighttime images, and one trained with half day and half night (see Fig \ref{fig:day_v_night}). We find that the models trained on only day and only night perform similarly on trans deer, and that the 50/50 split performs best on trans deer (highlighted region in Fig \ref{fig:day_v_night}). Training on day or night alone gives us a 20\% performance boost on trans deer, while training on both gives us a 40\% performance boost. This suggests that the day and night simulated images help the classifier in complementary ways: day helps with day images and night helps with night images. Performance on other classes is not strongly effected. Cis performance is quite noisy, and performs best with no added simulated data, see Fig. 2 in the main paper for further analysis.

\setlength{\tabcolsep}{4pt}
\begin{table}[h!]
\begin{center}
\caption{Error with and without the 44 real deer examples when adding 100K simulated deer images. Error is computed as in Table \ref{table:architecture}.}
\label{table:no_real_deer}
\begin{tabular}{l|cc|cc|}
\cline{2-5}
&  \multicolumn{2}{|c|}{Cis Test} & \multicolumn{2}{|c|}{Trans+ Test} \\
\hline
\multicolumn{1}{|c|}{Real Training Data}& Deer & Other & Deer & Other \\
\hline
\multicolumn{1}{|c|}{CCT train w/o deer} & 94.29 & 18.64 & 68.56 & 34.42 \\
\multicolumn{1}{|c|}{CCT train w/ deer} & 52.14 & 10.91 & 44.05 & 30.47  \\
\multicolumn{1}{|c|}{\% decrease from real deer} & 44.7 & 41.5 & 35.7 & 11.5  \\
\hline
\end{tabular}
\end{center}
\end{table}

\subsection{Investigating the effect of adding simulated data for a common class}\label{wolves}
In order to investigate how added simulated data might effect a common class, as opposed to a rare one, we created ``coyote" simulated data with TrapCam-Unity, using rendered models of wolves as a proxy for coyotes. Off-the-shelf, high-quality wolf models were more widely available, and wolves and coyotes are visually very similar (see Fig.\ref{fig:coyote_v_wolf}).  This is a coarse-grained experiment, and it remains to be seen what would happen if simulated data from two visually similar classes (\eg wolves and coyotes) was added at the same time. 

We find that adding simulated ``coyote" data improves trans+ coyote performance slightly, while cis coyote performance remains the same. Unsurprisingly, for the deer class (which has few training examples) adding a large amount of simulated coyote data harms both cis and trans+ deer performance.
\begin{figure}[t]
\centering
\includegraphics[width=8.5cm]{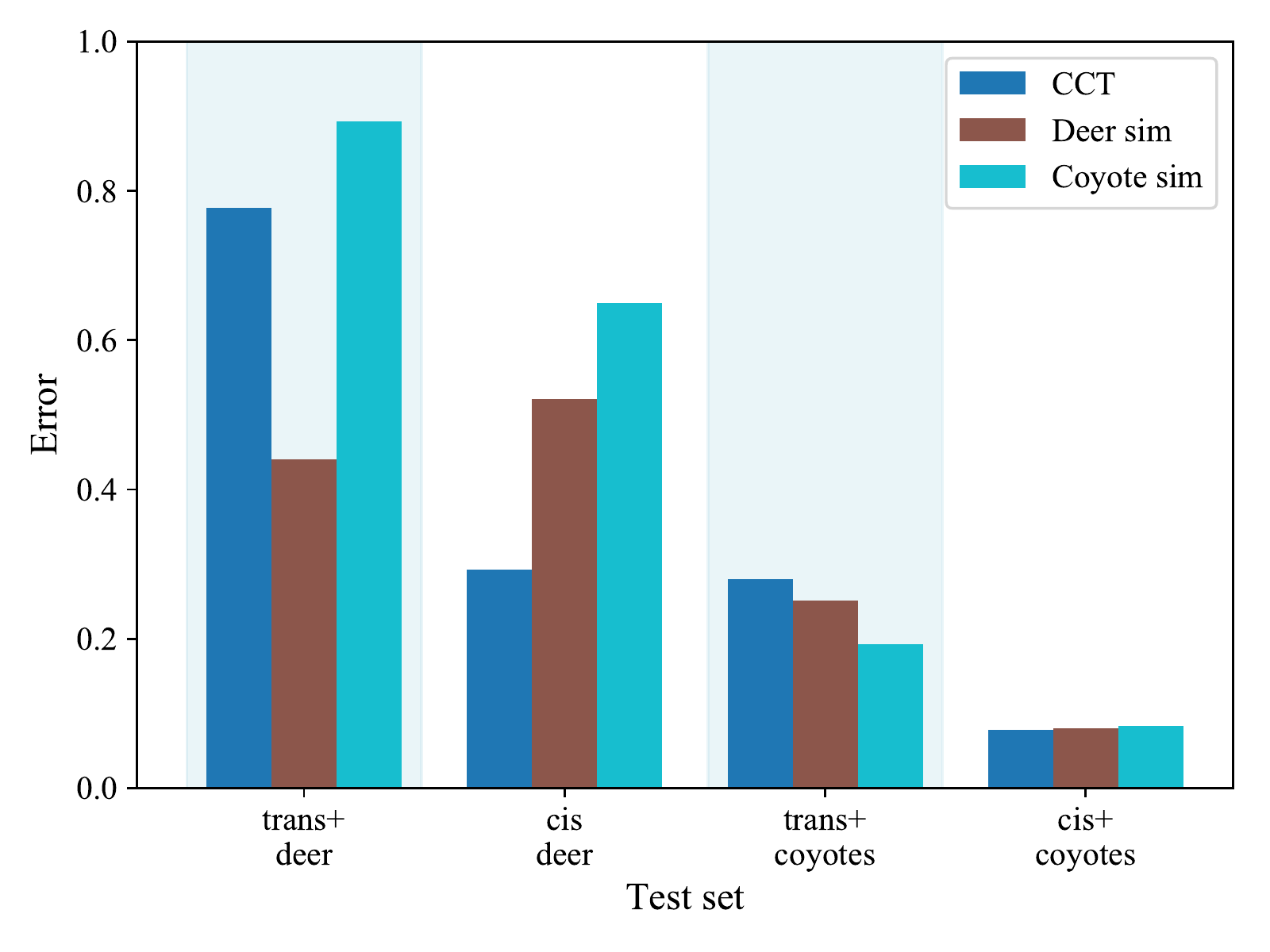}
\vspace{-15pt}
\setlength{\belowcaptionskip}{-12pt}
\caption{{\bf Error as a function of deer or coyote simulated images: 100K simulated images.} Error is calculated as in Fig. 4 in the main paper. Trans+ deer and coyote performance are highlighted.}
\label{fig:deer_v_}
\end{figure}

\begin{figure}[t]
\centering
\subfloat[Coyote \cite{coyote_im}]{\includegraphics[width=3.5cm]{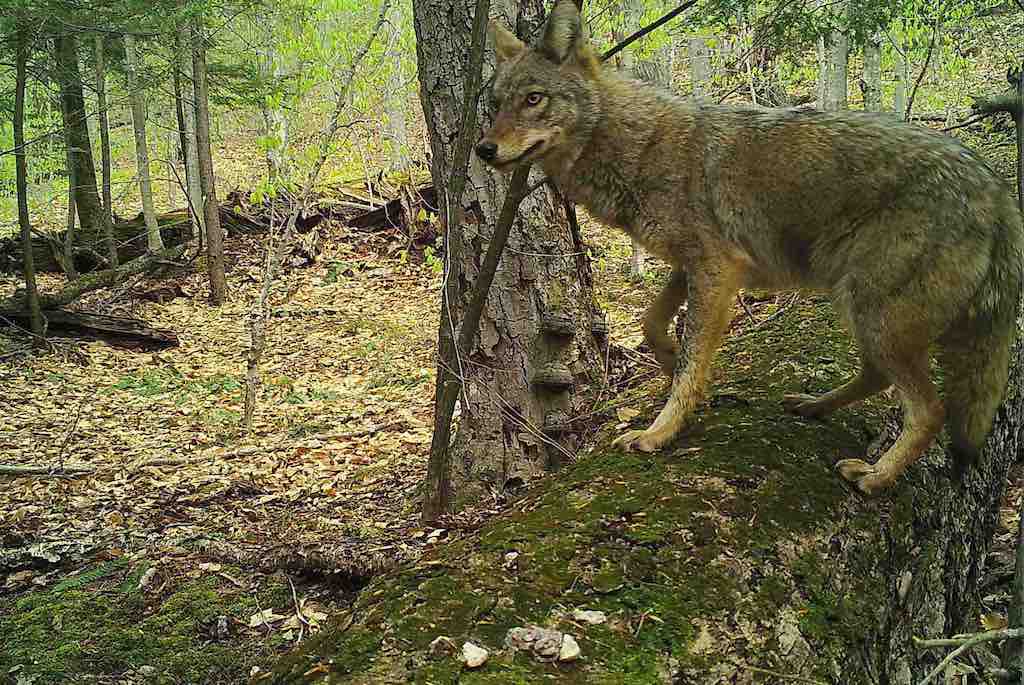}}\quad
\subfloat[Wolf \cite{wolf_im}]{\includegraphics[width=3.5cm]{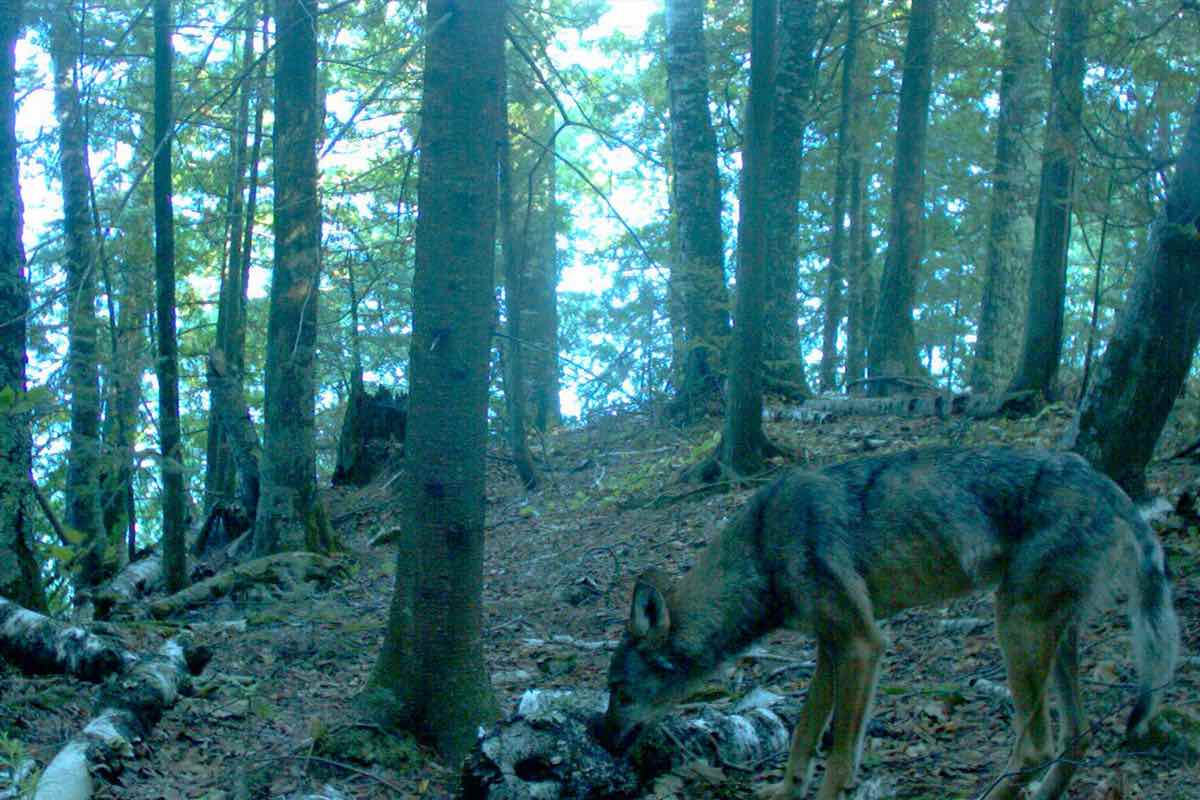}}
\vspace{-5pt}
\setlength{\belowcaptionskip}{-12pt}
\caption{{\bf Wolves and coyotes are visually similar.} }
\label{fig:coyote_v_wolf}
\end{figure}

\section{Creating Sim and Real on Empty Data}
\begin{figure}[h]
\centering
    \begin{tabular}{@{}ccc@{}}

    \subfloat[Simulated deer foreground]{\includegraphics[height=2.5cm]{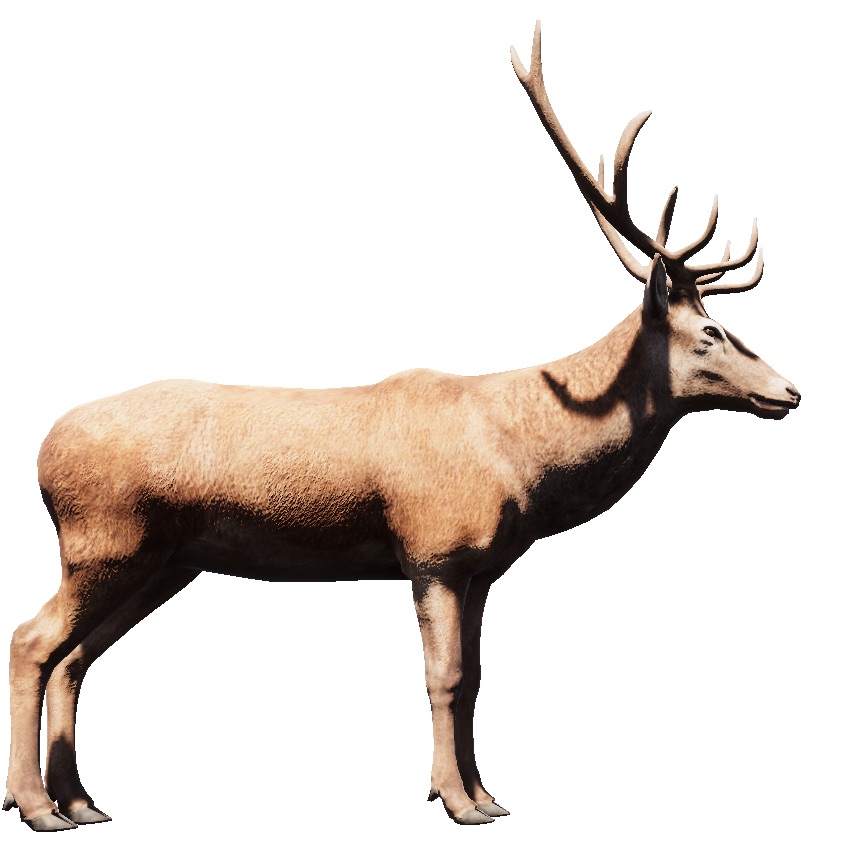}} &
    \subfloat[Cropped real deer foreground]{\includegraphics[height=2.5cm]{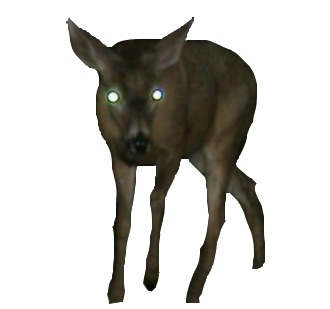}}\\
    
    \subfloat[Empty background from CCT]{\includegraphics[height=2.5cm]{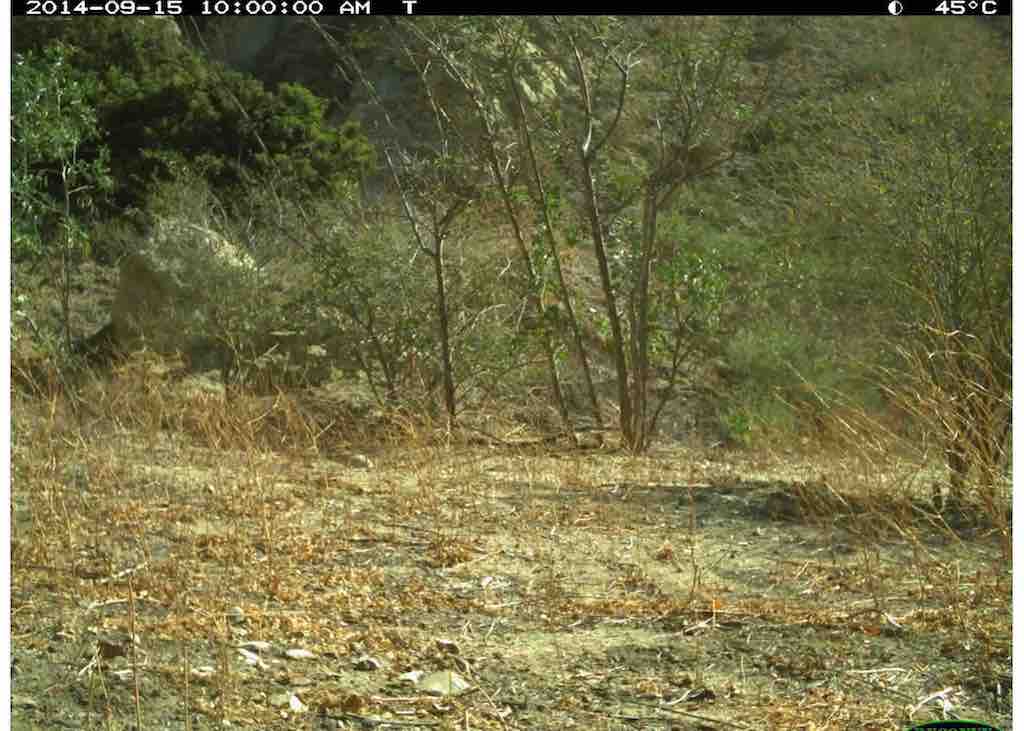}}&
    \subfloat[Empty background from CCT]{\includegraphics[height=2.5cm]{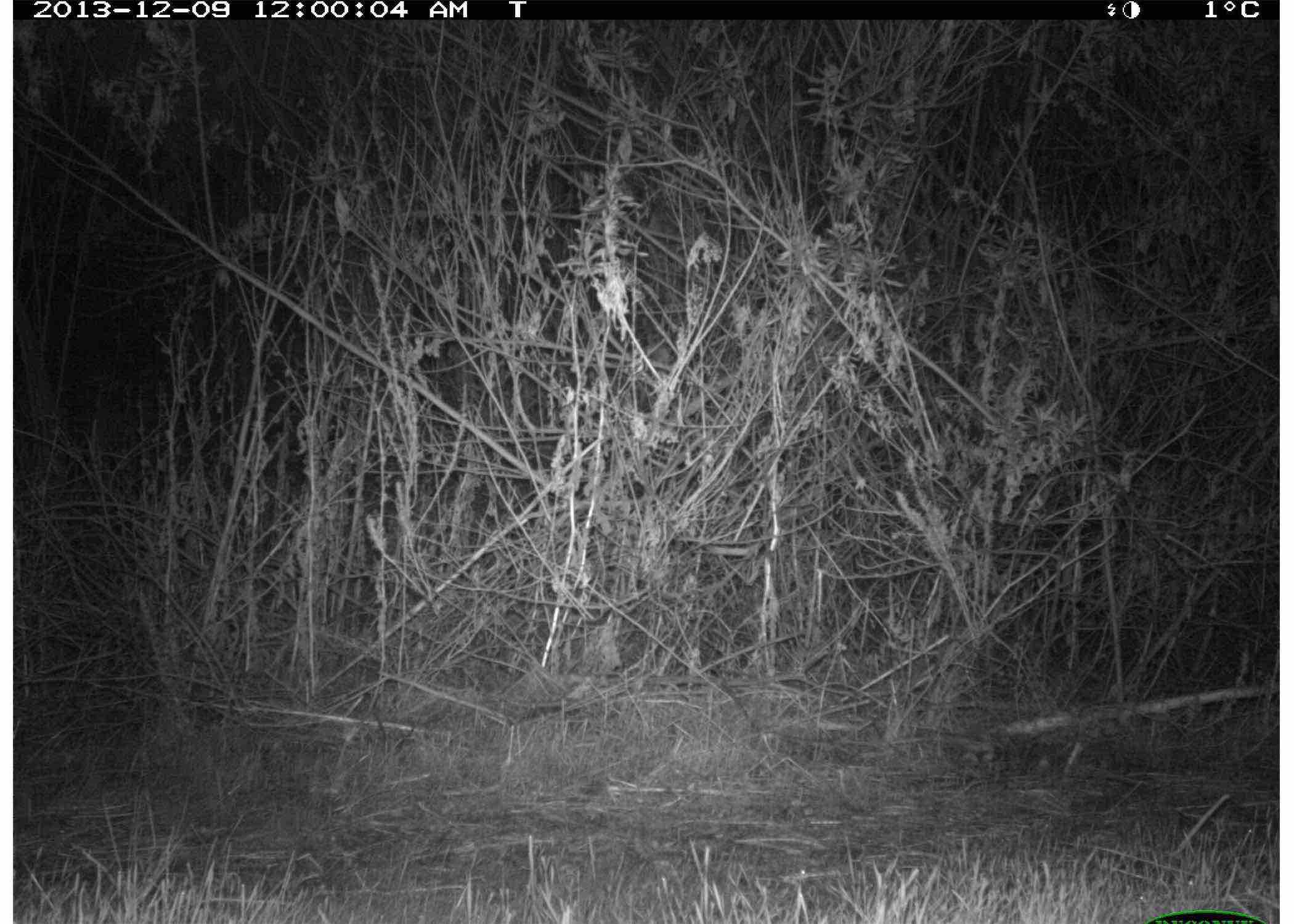}}\\

    \subfloat[Sim on empty overlay]{\includegraphics[height=2.5cm]{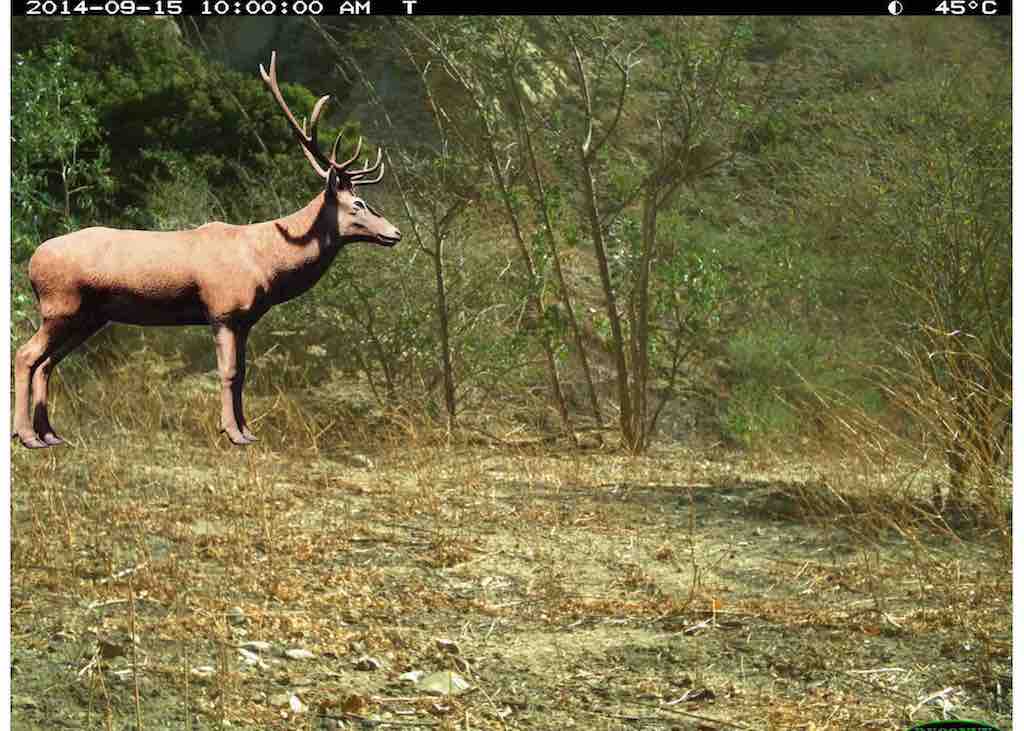}}&
    \subfloat[Real on empty overlay]{\includegraphics[height=2.5cm]{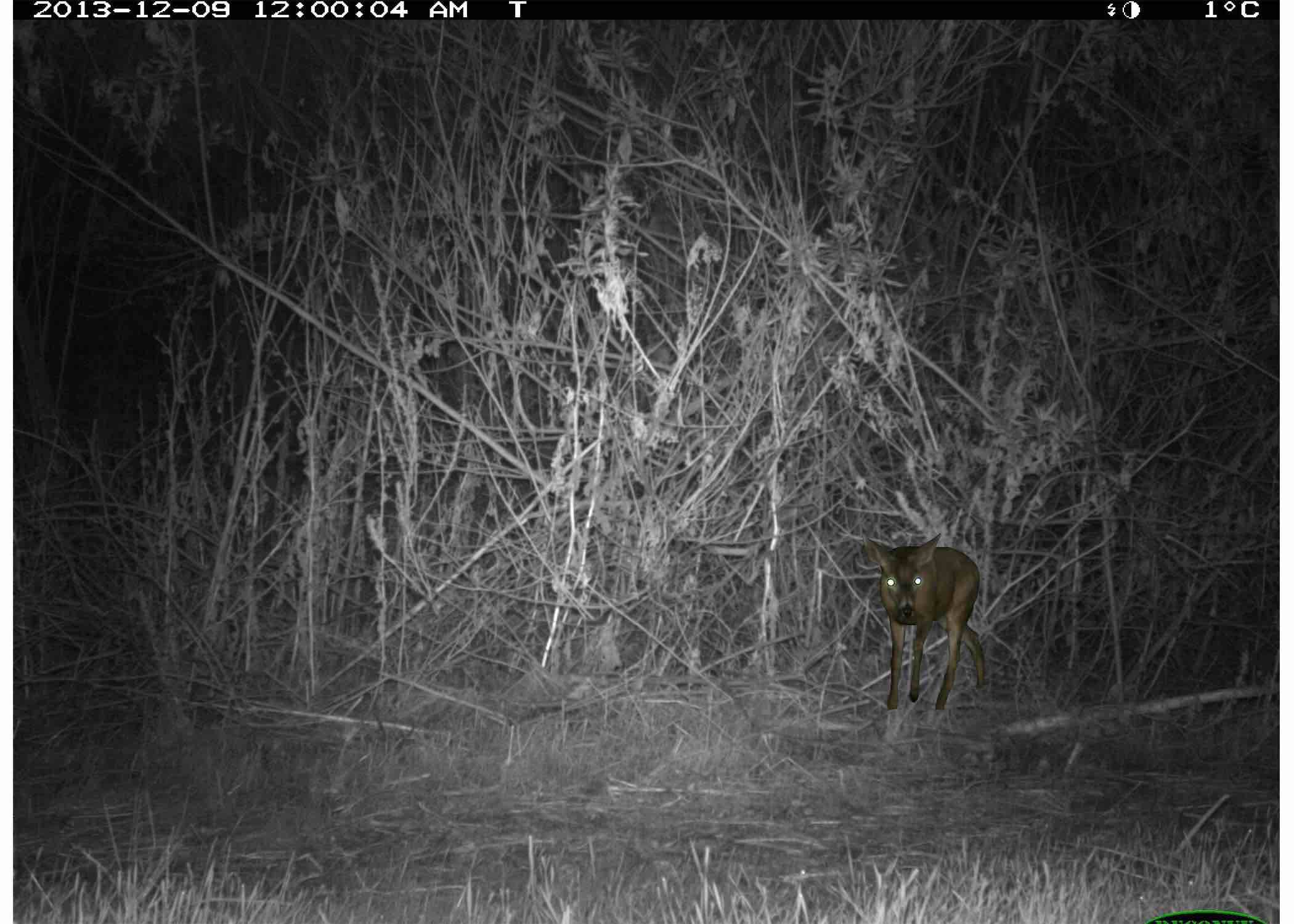}}\\
    \end{tabular}
    \caption{{\bf Sim and Real on Empty Generation.} (a),(c),(e) demonstrate the process of overlaying a simulated deer on top of an empty background image from the CCT dataset. (b),(d),(f) show the process of overlaying a cropped real deer on top of an empty background image from the CCT dataset.}
    \label{fig:sim_real_on_empty}
\end{figure}

Alternative to the full synthetic methods of data generation with AirSim and Unity, we generated synthetic images by overlaying either simulated deer or real cropped deer on real empty background images from the CCT dataset (see Fig. \ref{fig:sim_real_on_empty}). 

For the \textit{Sim on Empty} dataset generation, we posed either a stag or a doe deer from the GiM model set in front of a simulated camera in Unity. We randomized the animation, orientation in azimuth (0-360 degrees), position, direction of light orientation in azimuth (0-360 degrees), and elevation (20-90 degrees).

For the \textit{Real on Empty} dataset, we manually segmented and cropped out the 44 instances of deer from the CCT training set. Then we pasted the cropped deer foreground images on top of empty camera trap images in random locations.

\section{TrapCam-AirSim Details}

It took time and thought to derive the overall requirements for the AirSim TrapCam environment. With a sizable number of potential biomes globally, we narrowed the scope of what we intended to build to a SW United States environment similar to what is seen in the CCT data. Eventually we settled on a sub-alpine woodland scene that is readily found across most of the Western/ Southwest US. A major requirement and challenge was how to get the most data out of a relatively small, but detailed, area - this was key to the project without expanding the size of the area of interest. The overall intent was to leverage Microsoft AirSim's computer vision mode to move a pre-configured camera around the scene, providing varied background.

\begin{figure}[h]
\centering
    \begin{tabular}{@{}ccc@{}}

    \subfloat{\includegraphics[height=2.3cm]{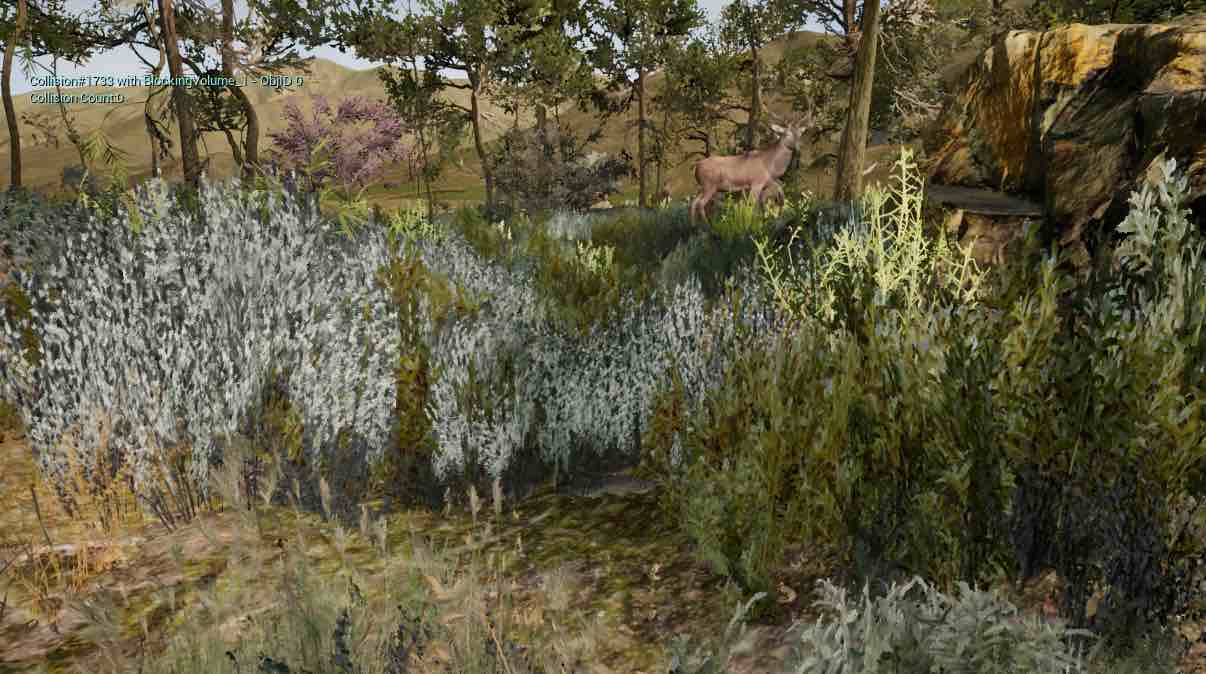}} &
    \subfloat{\includegraphics[height=2.3cm]{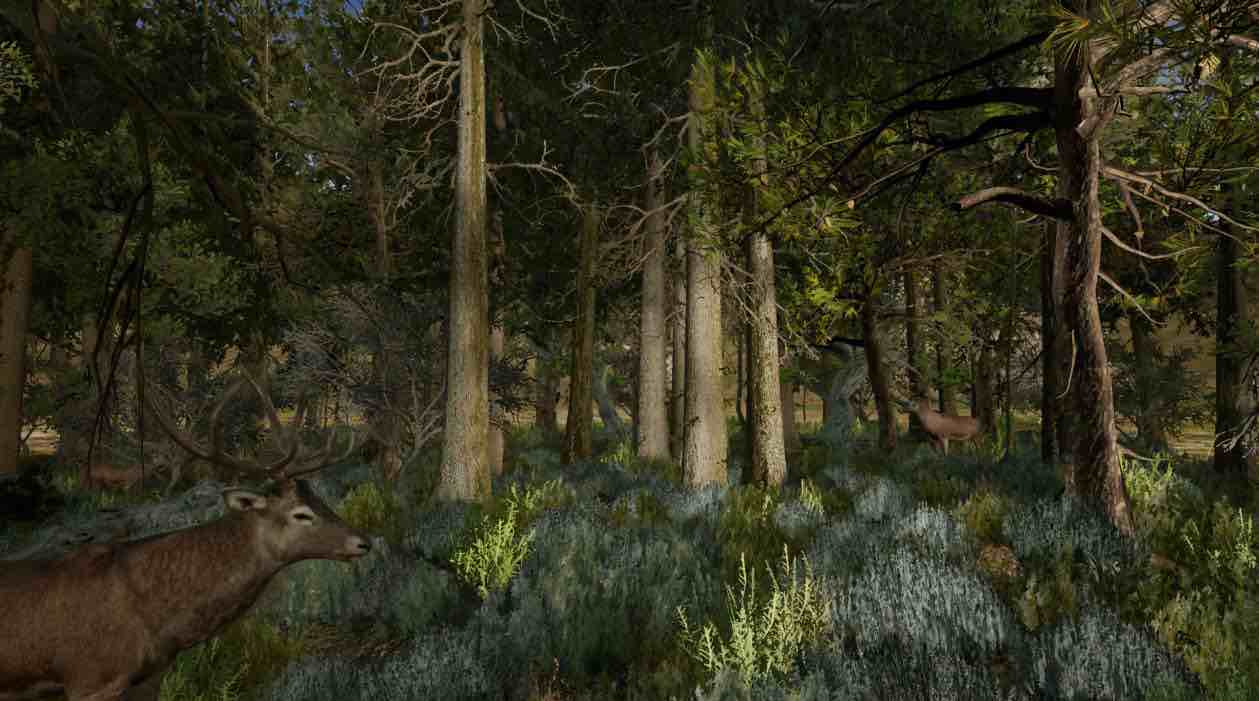}}\\
    \subfloat{\includegraphics[height=2.3cm]{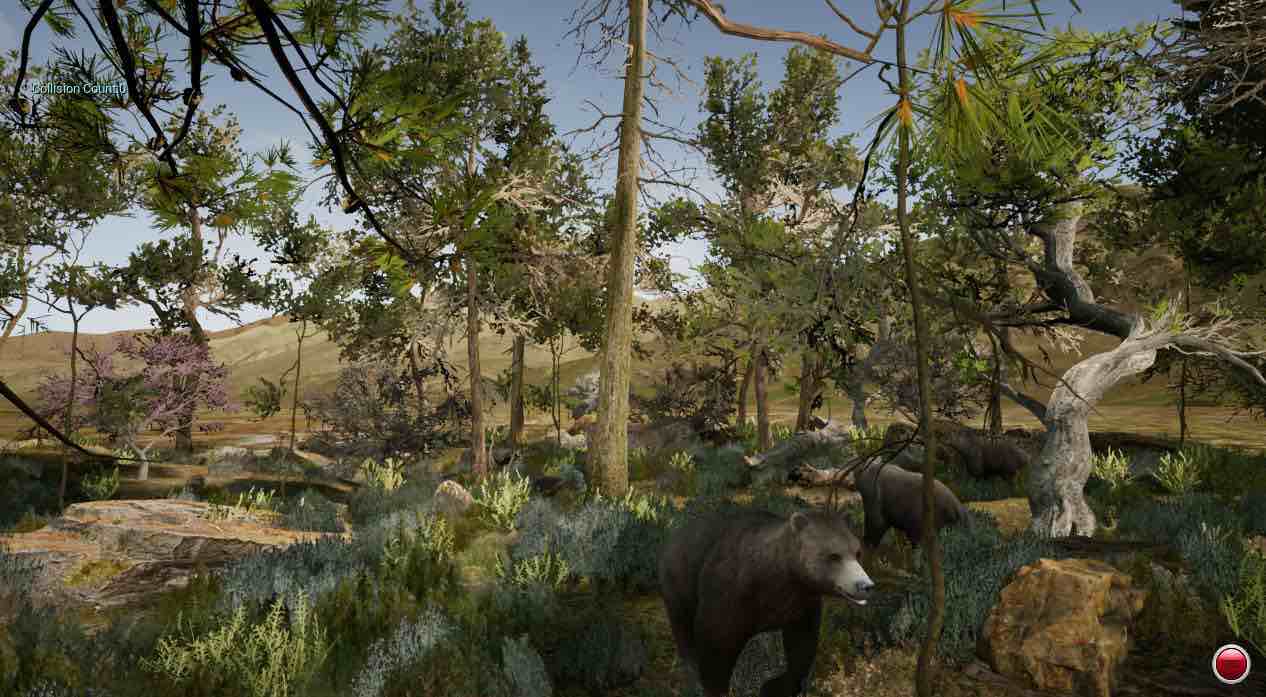}}&
    \subfloat{\includegraphics[height=2.3cm]{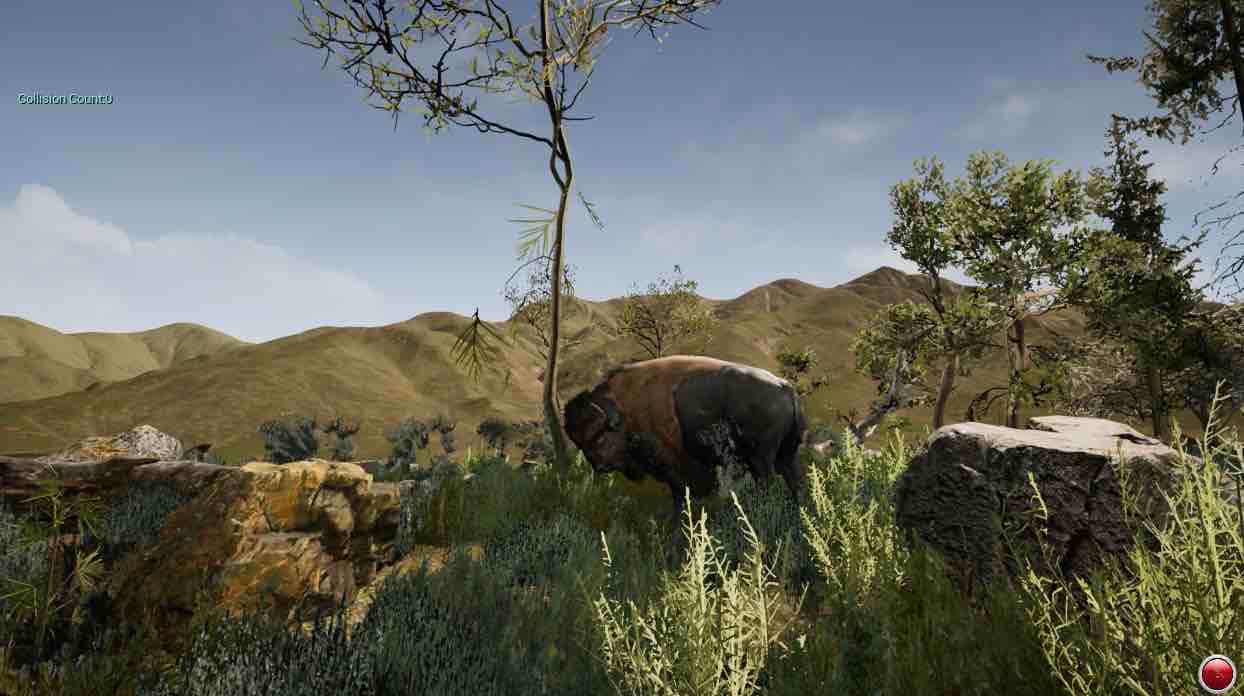}}\\
    \subfloat{\includegraphics[height=2.3cm]{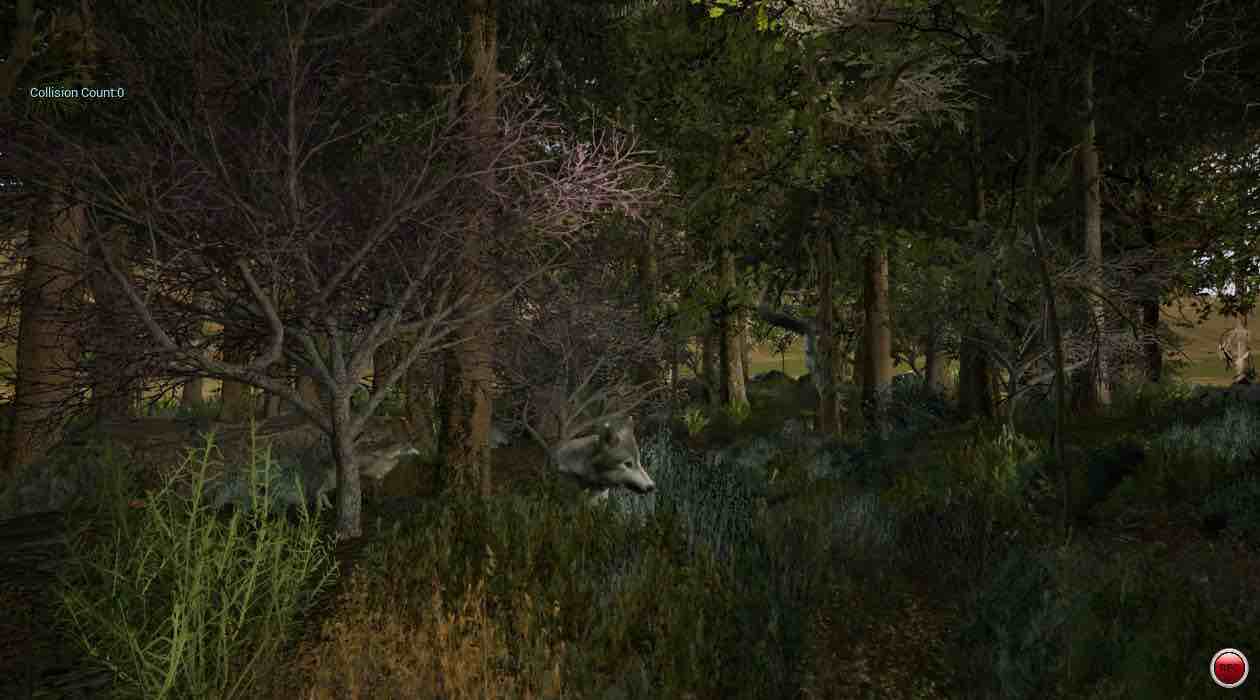}}&
    \subfloat{\includegraphics[height=2.3cm]{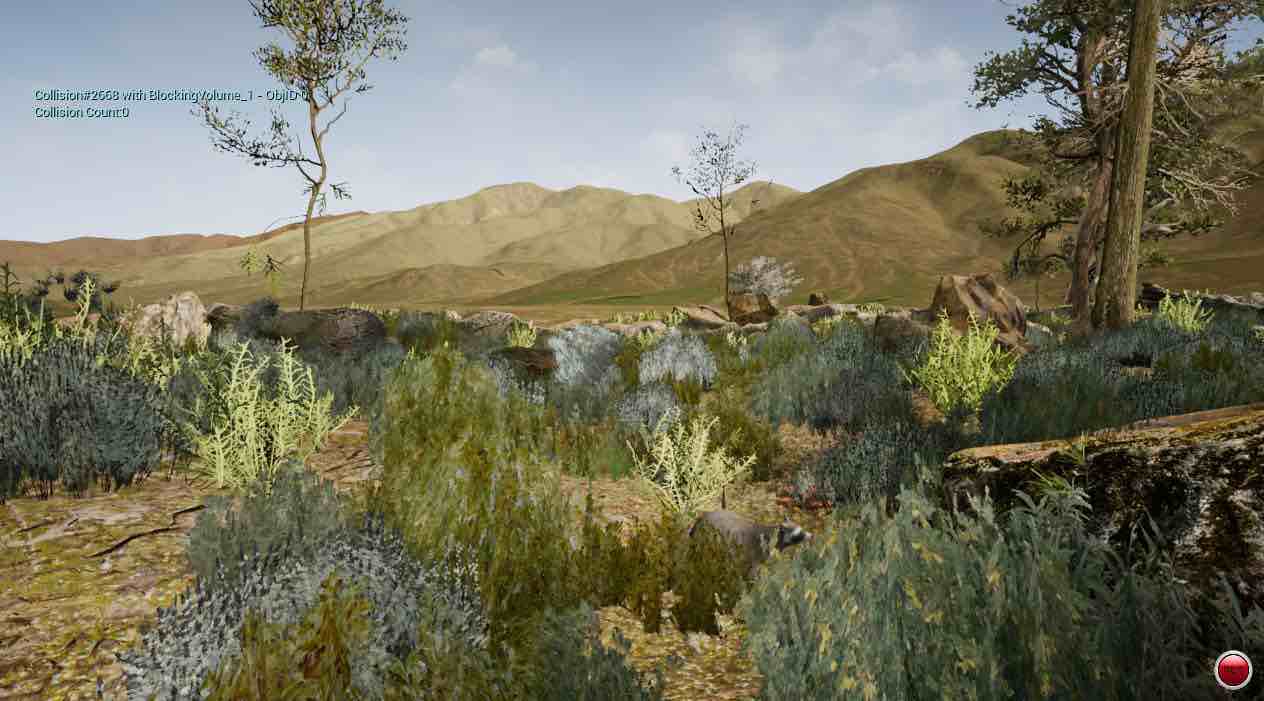}}\\
    \end{tabular}
    \caption{{\bf TrapCam-AirSim environment.} The TrapCam-Airsim envionment was designed to be modular and randomizeable, which allows a variety of biomes to be synthesized within a limited simulated area.}
    \label{fig:airsim_env}
\end{figure}

We used various off-the-shelf components such as an animal pack from Epic Studios \cite{epicstudios} (Animals Vol 01: Forest Animals by GiM \cite{gim}), background terrain from Unreal Marketplace \cite{UNREAL}, vegetation from SpeedTree \cite{speedtree}, and rocks/obstructions from Megascans \cite{megascans}. In other AirSim environments, the general scenery is fairly static with exception of particle effects (snow/rain/dust/etc). For this effort we wanted a method to vary the background, to replicate a variety of terrains within a single environment (see Fig.\ref{fig:airsim_env}). The actual area of the environment is small, at $50$ meters long, but the modularity allows many possible scenes to be constructed. The randomization was designed to facilitate artists by allowing them to make a list of different objects to randomize from. Those objects are prioritized based on their order on the list. The BiomeTerrain class generates them by tracing random areas across the field based on a global seed. If there's space available it spawns the desired object. There are a number of object types available in TrapCam-AirSim; animal type, rocks, logs, grasses, shrubs, trees, and each type can be varied by density and distribution. Additionally, we provide 9 GiM animal models: deer (doe/stag), wolf, fox, rat, spider, bear, raccoon, and buffalo. The doe model was created by removing the antlers from the stag model with Maya \cite{maya}, a common modeling tool. All animal objects were assigned segmentation IDs for efficient ground truth extraction. 

We created a simple UI to vary parameters, along with a command line API for parameter configuration. The UI was constructed with Unreal Motion Graphics (UMG) Widgets and allows for future flexibility for modifications, DPI resolutions and platforms. The main core functionalities were created with C++ for better performance as a parent class for data-only blueprints, which allows the technical artists to easily swap assets for different environments without re-compiling the C++ code.

We started the requirements and scoping in mid-August 2018 with a go-ahead approximately 6 September, and produced a working prototype two weeks later, with continued development and refining through mid-October. A second phase late in the year modified the camera system to include flash capability, and animals were updated to provide eye-shine, and the UI was modified to include variability for that eye-shine.

\section{TrapCam-Unity Details}
\label{unity}
\begin{figure}[t]
\centering
\begin{tabular}{@{}ccc@{}}
\subfloat[Models of deer]{
\includegraphics[width=6cm]{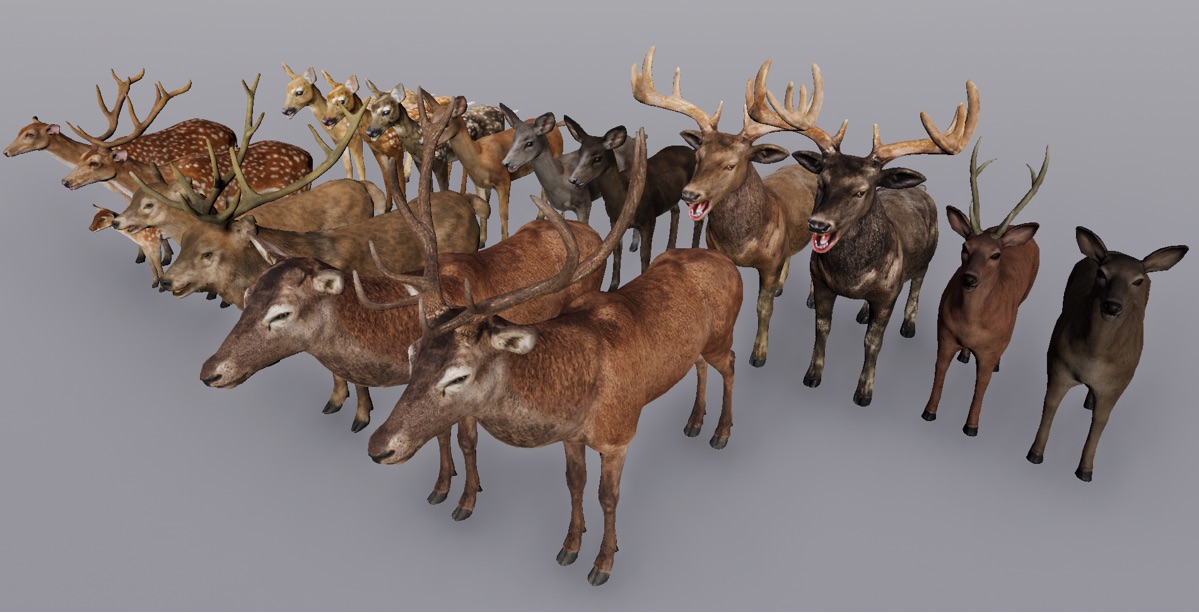}}\\
\subfloat[Models of wolves]{
\includegraphics[width=6cm]{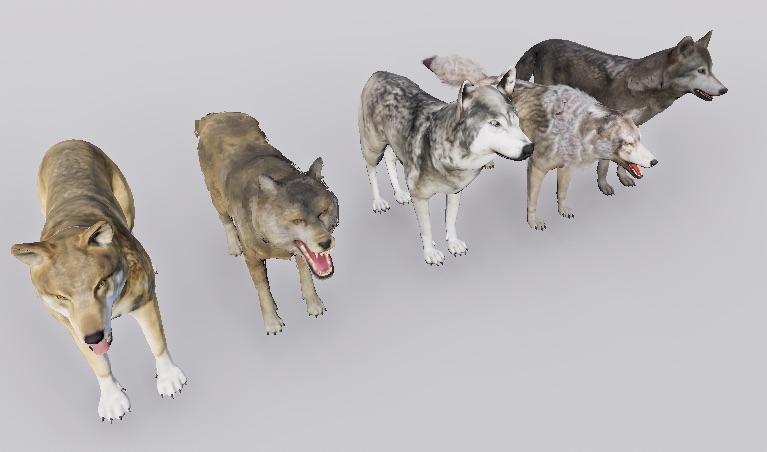}}
\end{tabular}
\vspace{-5pt}
\setlength{\belowcaptionskip}{-15pt}
\caption{{\bf Models of deer and wolves.} In TrapCam-Unity we used 17 different models of deer from 5 different artists and 5 models of wolves from 5 different artists. We used the wolf models as proxies for coyotes (see Section \ref{wolves}). Model details are available in Section \ref{unity}.}
\label{fig-models}
\end{figure}

\begin{figure}[h]
\centering
    \begin{tabular}{@{}ccc@{}}

    \subfloat{\includegraphics[height=2.3cm]{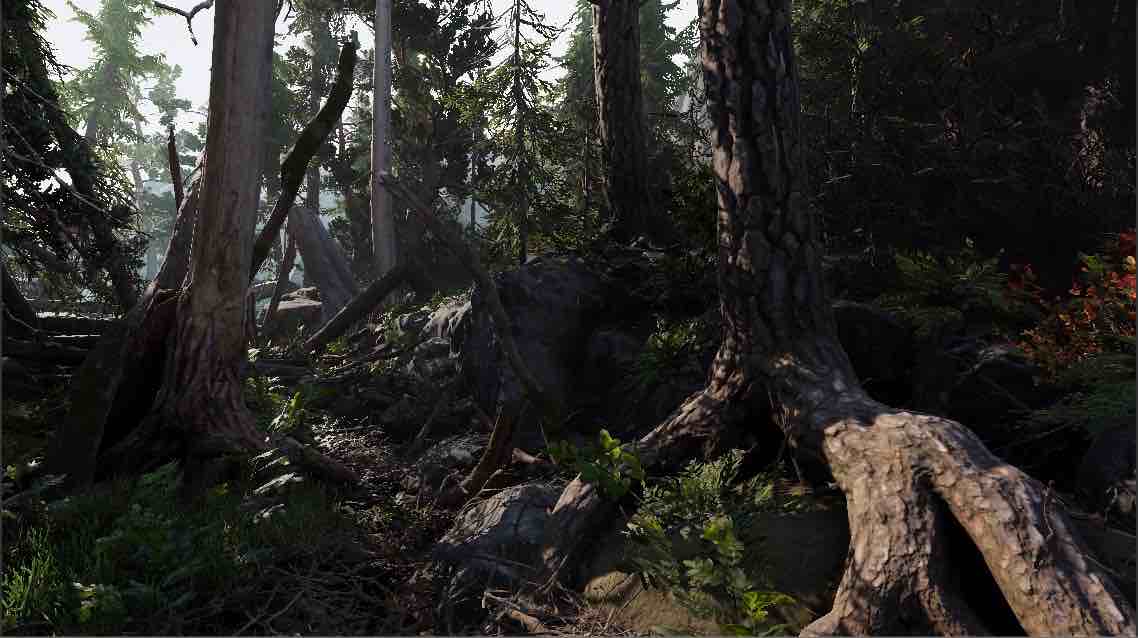}} &
    \subfloat{\includegraphics[height=2.3cm]{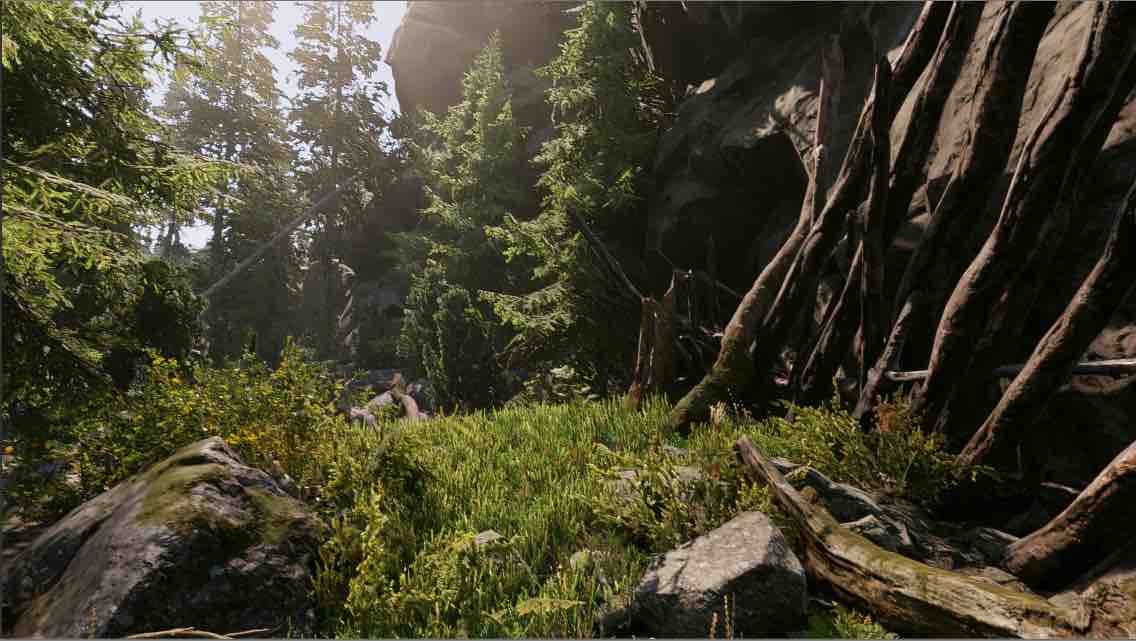}}\\
    \subfloat{\includegraphics[height=2.3cm]{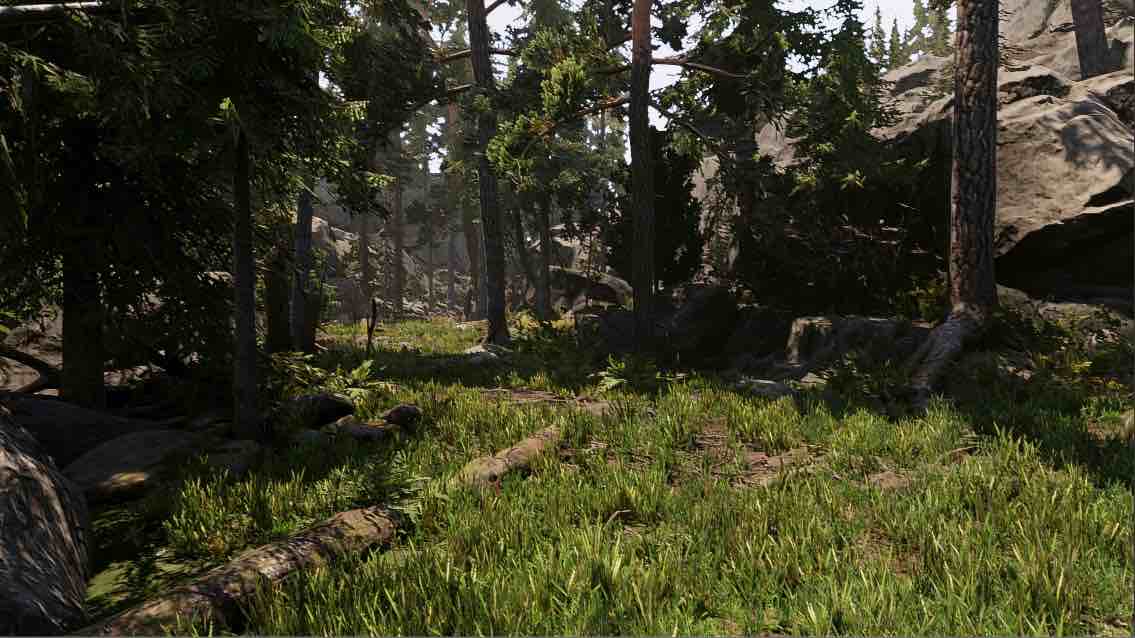}}&
    \subfloat{\includegraphics[height=2.3cm]{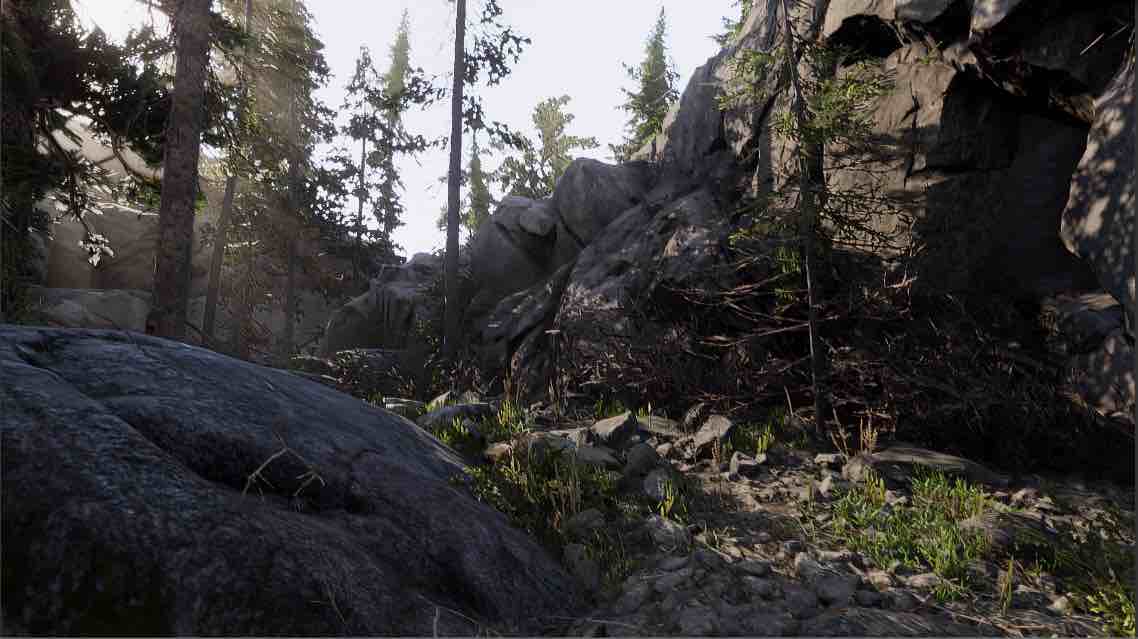}}\\
    \subfloat{\includegraphics[height=2.3cm]{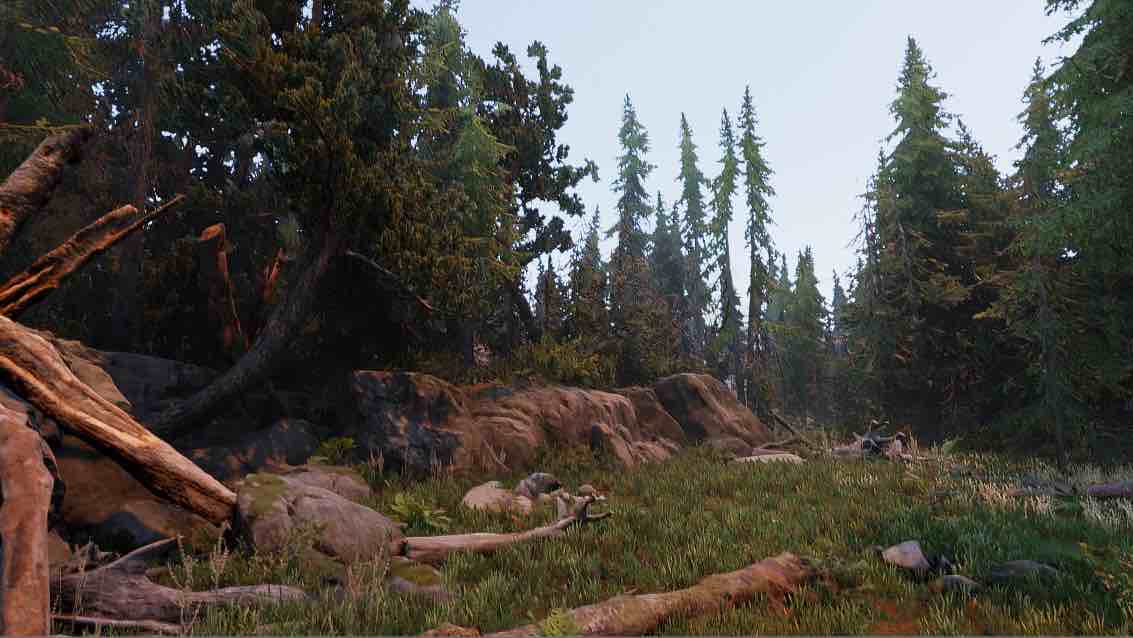}}&
    \subfloat{\includegraphics[height=2.3cm]{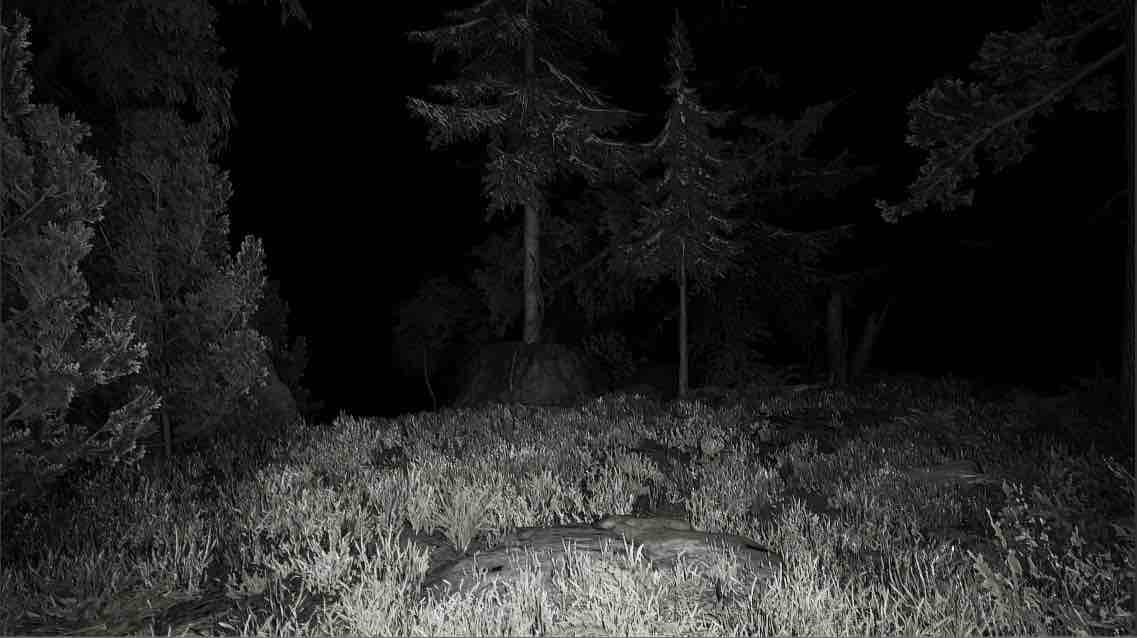}}\\
    \end{tabular}
    \caption{{\bf TrapCam-Unity environment.} The Book of The Dead environment is a large natural environment with diverse sub regions.}
    \label{fig:bod_env}
\end{figure}

The ``Book of The Dead" environment~\cite{bod} we use is published for free by Unity. As shown in Fig.\ref{fig:bod_env}, the near-photorealistic environment simulates a large patch of forest in a valley with volumetric grass, a variety of high definition trees, logs, and bushes, as well as rocks and terrain. The environment is a irregular area of roughly 20,000 $m^2$. It runs on a desktop PC in real time and enables us to generate large amounts of images efficiently. 

\begin{figure}[h]
    \centering
    \includegraphics[width=7cm]{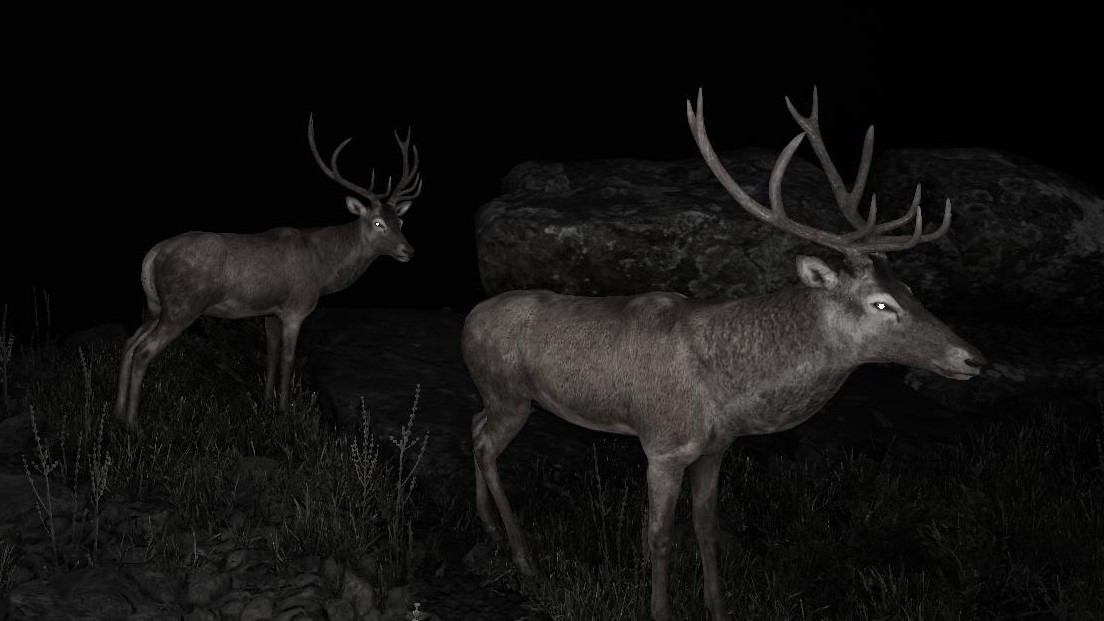}\par
    \caption{\bf Example of eyeshine simulation.}
    \label{fig:eyeshine}
\end{figure}

To create daytime images we varied the orientation of the simulated sun in both azimuth and elevation. To create images taken at night we created a spotlight attached to the simulated camera to simulate a white-light or IR flash and qualitatively match the low color saturation of the night time images. To simulate animals' eyeshine (a result of the reflection of camera flash from the tapetum lucidum), we placed small reflective balls on top of the eyes of model animals (see Fig.\ref{fig:eyeshine}).

For deer simulation, we used 17 animated deer models from 5 publishers on Unity (GiM\cite{gim_unity}, 4toon\cite{4toon_unity}, Protofactor\cite{protofactor_unity}, Red Deer\cite{radik_bilalov_unity}, Janpec\cite{janpec_unity}). For coyote simulation, we used 5 models from 5 publishers (GiM\cite{gim_unity}, 4toon\cite{4toon_unity}, Protofactor\cite{protofactor_unity}, Janpec\cite{janpec_unity}, WDallgraphics\cite{wdallgraphics}). We created the GiM doe model by removing the antlers of the GiM stag model with Blender\cite{blender}. For each of the animated models, we included an animation controller that contains several animation clips ranging from commonly seen behavior episodes like walking and eating, to rare occurrences like attacking and sleeping. During dataset generation, we randomly picked a clip for each instance of animals and freeze it at a random time point, then we move the cameras around to sample a static scene with animals and environment. 

We had ~300 seed locations and randomly placed animals in the vicinity of a subset of the seed locations. This process was repeated multiple times to simulate animals in random locations within the environment. A similar random placement process was used to determine the locations of the cameras. All images generated are in full HD resolution (1980 x 1080).

For ground truth generation, we turned off the lighting and rendered each instance of the animal in a unique color by replacing the original animal shader with an unlit shader. We then used customized python scripts to extract animal bounding boxes by extracting pixels with these unique colors. 

\end{document}